\documentclass[10pt,twocolumn,letterpaper]{article}

\usepackage{cvpr}              

\usepackage{comment}
\usepackage{graphicx}
\usepackage{amsmath}
\usepackage{amssymb}
\usepackage{booktabs}
\usepackage{caption}
\usepackage{subcaption}
\usepackage{graphics}
\usepackage{comment}
\usepackage{enumitem}
\usepackage{multirow}
\usepackage[table,dvipsnames]{xcolor}

\usepackage{amsmath}
\usepackage{bm}
\usepackage{bbm}

\usepackage{algorithm}
\usepackage{algorithmic}

\newcommand{\va}{\bm{a}}

\newcommand{\vy}{\bm{y}}

\newcommand{\mM}{\bm{M}}

\newcommand{\calW}{\mathcal{W}}

\newcommand{\norm}[1]{\lVert #1 \rVert}

\usepackage[pagebackref,breaklinks,colorlinks]{hyperref}

\usepackage[capitalize]{cleveref}
\crefname{section}{Sec.}{Secs.}
\Crefname{section}{Section}{Sections}
\Crefname{table}{Table}{Tables}
\crefname{table}{Tab.}{Tabs.}

\def\cvprPaperID{11855} 
\def\confName{CVPR}
\def\confYear{2023}

\newtoggle{comments}
\toggletrue{comments}
\iftoggle{comments}{%
    \newcommand{\achal}[1]{{\leavevmode\color{orange}[Achal: #1]}}
    \newcommand{\austin}[1]{{\leavevmode\color{magenta}[Austin: #1]}}
    \newcommand{\mariya}[1]{{\leavevmode\color{red}[Mariya: #1]}}
    \newcommand{\arjun}[1]{{\leavevmode\color{purple}[Arjun: #1]}}
}{%
    \newcommand{\achal}[1]{}
    \newcommand{\austin}[1]{}
    \newcommand{\mariya}[1]{}
    \newcommand{\arjun}[1]{}
}

\newcommand{\atmostlinewidth}{
  \ifdim\width>\columnwidth
    \columnwidth
  \else
    \width
  \fi}

\begin{document}

\title{HandsOff: Labeled Dataset Generation With\\ No Additional Human Annotations}

\author{Austin Xu\thanks{Work done as an intern at Amazon. \texttt{axu@gatech.edu}}\\
\normalsize{Georgia Institute of Technology}
\and
Mariya I. Vasileva\\
\normalsize{Amazon AWS}
\and 
Achal Dave\thanks{Work done while at Amazon}\\
\normalsize{Toyota Research Institute}
\and
Arjun Seshadri\\
\normalsize{Amazon Style}
}

\maketitle

\begin{abstract}
 Recent work leverages the expressive power of generative adversarial networks (GANs) to generate labeled synthetic datasets. These dataset generation methods often require new annotations of synthetic images, which forces practitioners to seek out annotators, curate a set of synthetic images, and ensure the quality of generated labels. We introduce the HandsOff framework, a technique capable of producing an unlimited number of synthetic images and corresponding labels after being trained on less than 50 pre-existing labeled images. Our framework avoids the practical drawbacks of prior work by unifying the field of GAN inversion with dataset generation. We generate datasets with rich pixel-wise labels in multiple challenging domains such as faces, cars, full-body human poses, and urban driving scenes. Our method achieves state-of-the-art performance in semantic segmentation, keypoint detection, and depth estimation compared to prior dataset generation approaches and transfer learning baselines. We additionally showcase its ability to address broad challenges in model development which stem from fixed, hand-annotated datasets, such as the long-tail problem in semantic segmentation. Project page: \url{austinxu87.github.io/handsoff}.
\end{abstract}

\section{Introduction}\label{sec:intro}
The strong empirical performance of machine learning (ML) models has been enabled, in large part, by vast quantities of labeled data. The traditional machine learning paradigm, where models are trained with large amounts of \textit{human labeled} data, is typically bottlenecked by the significant monetary, time, and infrastructure investments needed to obtain said labels. This problem is further exacerbated when the data itself is difficult to collect. For example, collecting images of urban driving scenes requires physical car infrastructure, human drivers, and compliance with relevant government regulations. 

Finally, collecting real labeled data can often lead to imbalanced datasets that are unrepresentative of the overall data distribution. 
For example, in \textit{long-tail settings}, the data used to train a model often does not contain rare, yet crucial edge cases \cite{zhang2021deep}.

\begin{figure}
    \centering
    \includegraphics[width=0.88\columnwidth]{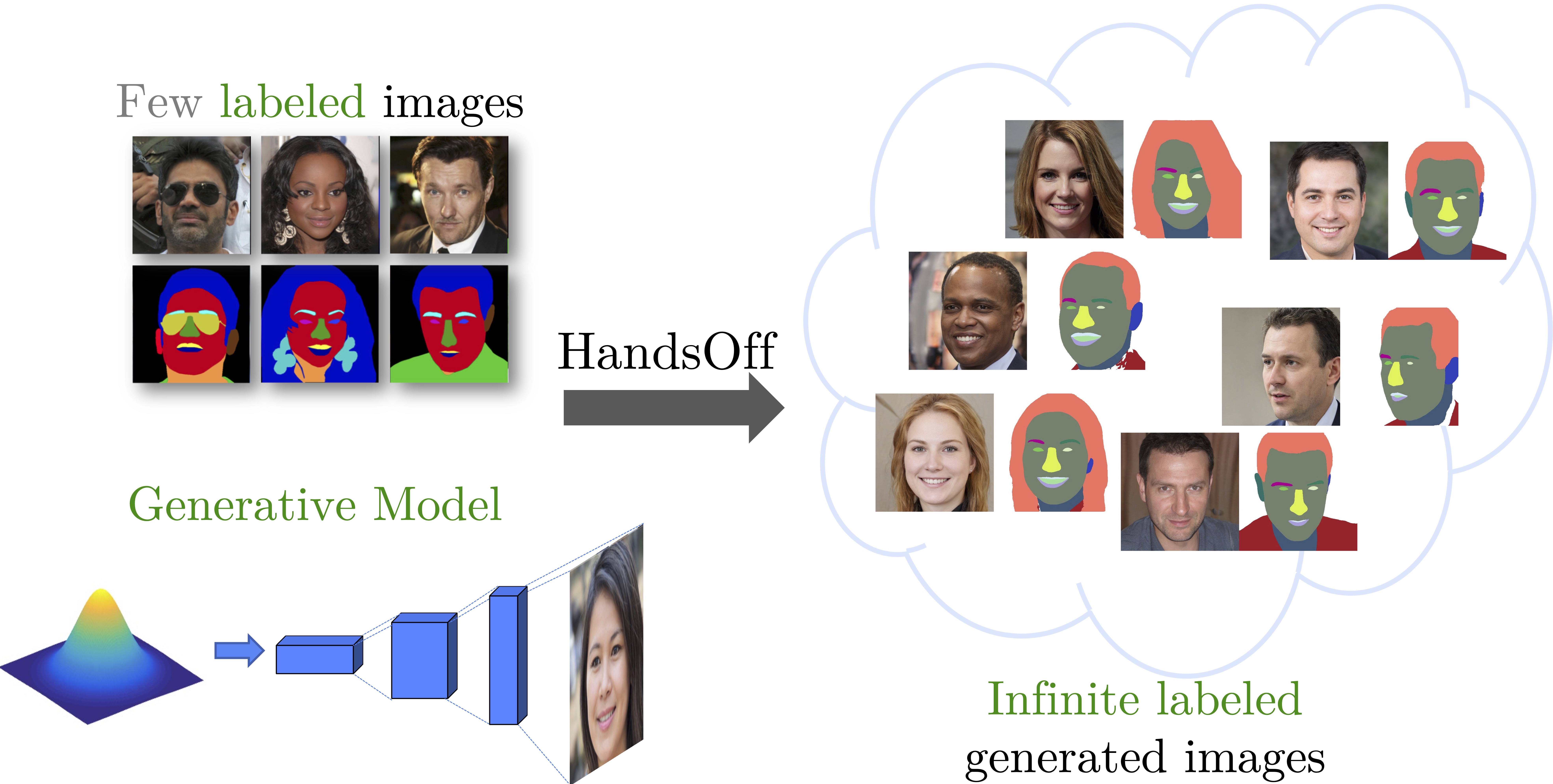}
    \caption{\small{The HandsOff framework uses a small number of existing labeled images and a generative model to produce \textbf{infinitely} many labeled images.}}
    \label{fig:high_level}
\end{figure}

These limitations make collecting ever increasing amounts of hand labeled data unsustainable.
We advocate for a shift away from the standard paradigm towards a world where training data comes from an \textit{infinite collection} of automatically generated labeled images. 
Such a dataset generation approach can allow ML practitioners to \textit{synthesize} datasets in a \textit{controlled} manner, unlocking new model development paradigms such as controlling the quality of generated labels and mitigating the long-tail problem.

In this work, we propose HandsOff, a generative adversarial network (GAN) based dataset generation framework.
HandsOff is trained on a small number of \textit{existing} labeled images and capable of producing an infinite set of synthetic images with corresponding labels (Fig.~\ref{fig:high_level}).

To do so, we unify concepts from two disparate fields: dataset generation and GAN inversion.
While the former channels the expressive power of GANs to dream new ideas in the form of images, the latter connects those dreams to the knowledge captured in annotations. In this way, our work brings together what it means to dream and what it means to know. Concretely, our paper makes the following contributions:
\begin{enumerate}
    \item We propose a novel dataset generating framework, called HandsOff, which unifies the fields of dataset generation and GAN inversion.
    While prior methods for dataset generation~\cite{zhang21datasetgan} require new human annotations on synthetically generated images, HandsOff uses GAN inversion to train on existing labeled datasets, eliminating the need for human annotations. With $\leq 50$ real labeled images, HandsOff is capable of producing high quality image-label pairs (Sec.~\ref{sec:framework}). 
    \item We demonstrate the HandsOff framework's ability to generate semantic segmentation masks, keypoint heatmaps, and depth maps across several challenging domains (faces, cars, full body fashion poses, and urban driving scenes) by evaluating performance of a downstream task trained on our synthetic data (Sec.~\ref{sec:exp:vis}, ~\ref{sec:exp:segmentation}, and~\ref{sec:exp:kp_depth}).
    \item We show that HandsOff is capable of mitigating the effects of the long-tail in semantic segmentation tasks. By modifying the distribution of the training data, HandsOff is capable of producing datasets that, when used to train a downstream task, dramatically improve performance in detecting long-tail parts (Sec.~\ref{sec:exp:long_tail}).
\end{enumerate}

\section{Related work}\label{sec:related_work}

\begin{figure*}[t]
    \centering
    \includegraphics[width=0.8\textwidth]{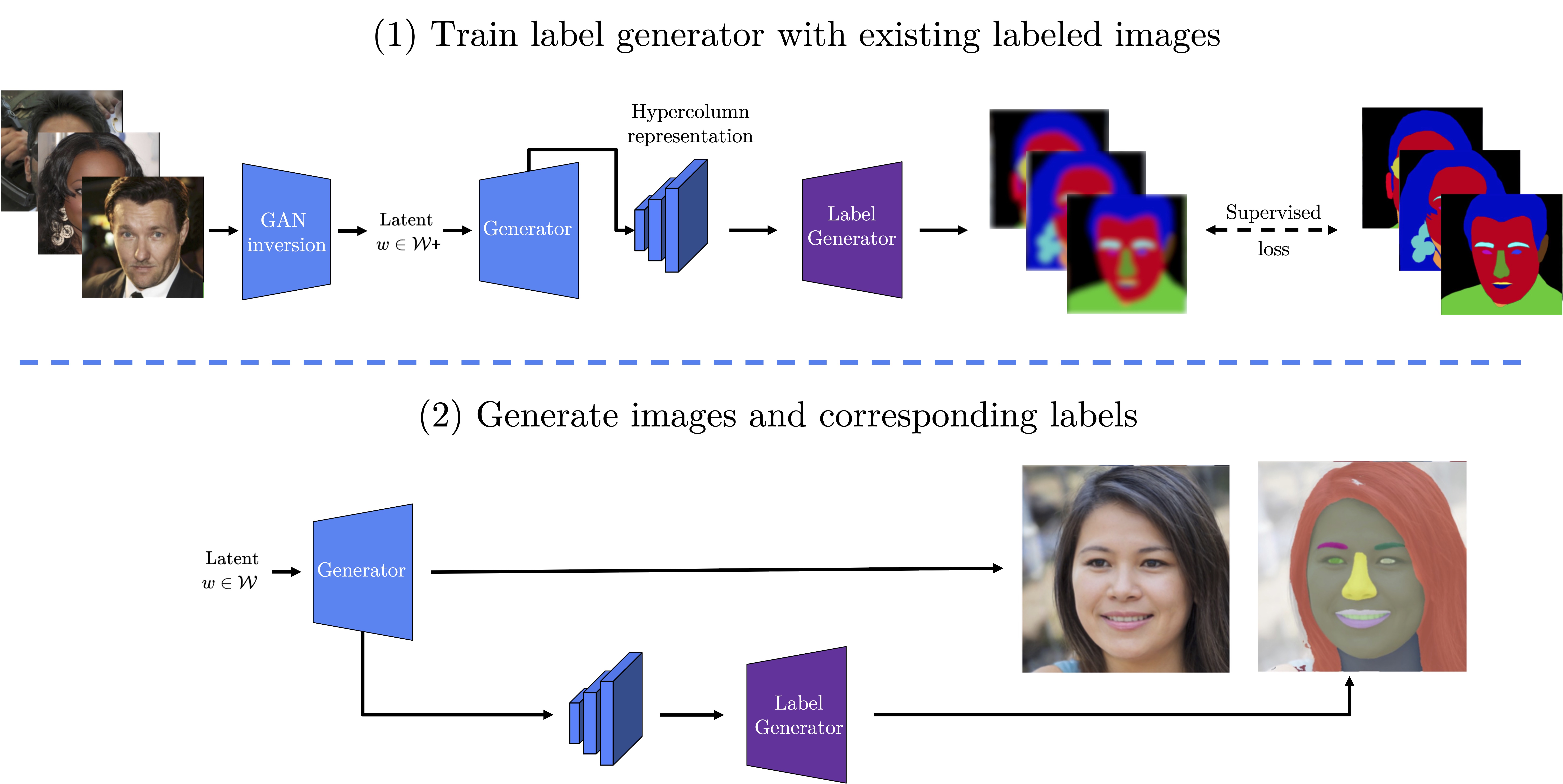}
    \caption{\small{The HandsOff framework. (Top) GAN inversion is used to obtain training image latent codes, which are then used to form hypercolumn representations. The label generator is then trained with the hypercolumn representations and original labels. (Bottom) To generate datasets, the trained label generator is used in conjunction with a StyleGAN2 generator to produce image-label pairs.}}
    \label{fig:framework}
\end{figure*}

Our work is built on GANs \cite{goodfellow2014generative}, which consist of a generator that synthesizes new images, and a discriminator that discerns between real and generated images. Recent advances in GANs \cite{brock2018large, karras2018progressive, karras2019style, karras2020analyzing, karras2020training, karras2021alias} have demonstrated an ability to generate highly realistic images in numerous domains. We utilize the popular StyleGAN2 architecture \cite{karras2020analyzing}, which synthesizes images by passing randomly sampled inputs through a series of \textit{style blocks}. Remarkably, StyleGAN2's $\calW$ and $\calW+$ latent spaces form rich representations of images in a disentangled manner \cite{xia2022gan,shen2020interpreting, abdal2019image2stylegan, abdal2020image2stylegan++}, which can be utilized to edit complex semantic attributes in generated images \cite{shen2020interpreting, menon2020pulse,harkonen2020ganspace, ling2021editgan, alaluf2021restyle, alaluf2022hyperstyle}. The ability to identify semantically meaningful parts of generated images in the latent representation suggests that it could be used to generate pixel-level labels. This capability, coupled with GANs' ability to generate troves of high quality images, serves as the basis for generating synthetic image \textit{datasets} \cite{zhang21datasetgan, semanticGAN, li2022bigdatasetgan, abdal2021labels4free}. 

We build upon DatasetGAN \cite{zhang21datasetgan}, which trains a label generator using representations of an image formed from the GAN latent code. DatasetGAN requires \textit{human annotation of GAN generated images}, which burdens a practitioner to seek out annotations for every new domain of interest. In addition to labeling, users also must actively \textit{curate} images to label to ensure diverse semantic feature coverage and avoid GAN created artifacts. Furthermore, should the labeling scheme change and render the original labels obsolete, then additional annotations are again required. Acquiring additional labels is especially contrived when a large of number of quality human annotated images already exist. A framework that leverages these \textit{real} preexisting labeled images would circumvent all of these drawbacks. EditGAN \cite{ling2021editgan}, a follow-on contribution to DatasetGAN, utilizes encoder-based reconstructions to perform image editing. BigDatasetGAN\cite{li2022bigdatasetgan} exploits the pre-trained encoder of VQGAN\cite{esser2021taming} to utilize existing labeled \textit{synthetic} images. In contrast, our approach links latents of labeled \textit{real} images to their labels by employing GAN inversion, the process of mapping a real image to the latent space of a GAN. 

The myriad of inversion techniques range from encoder-based approaches \cite{tov2021designing, alaluf2021restyle, richardson2021encoding, wang2022high}, which utilize trained encoders to map images directly to the latent space, to optimization-based approaches \cite{collins2020editing, abdal2019image2stylegan, abdal2020image2stylegan++}, which directly optimize a similarity loss (e.g., LPIPS \cite{zhang2018unreasonable}) to obtain latents. Some methods modify generator weights to increase image reconstruction quality \cite{roich2022pivotal, alaluf2022hyperstyle, abdal2020image2stylegan++}. Our work exclusively uses inversion methods that do not modify the generator, since the generator must remain unperturbed to generate new images from the original data distribution. We invert images to the $\calW+$ space, which is more expressive than the $\calW$ space and leads to higher quality reconstructions \cite{xia2022gan}. 

\section{The HandsOff framework}\label{sec:framework}
\begin{figure*}[t]
    \centering
    \includegraphics[width=0.9\textwidth]{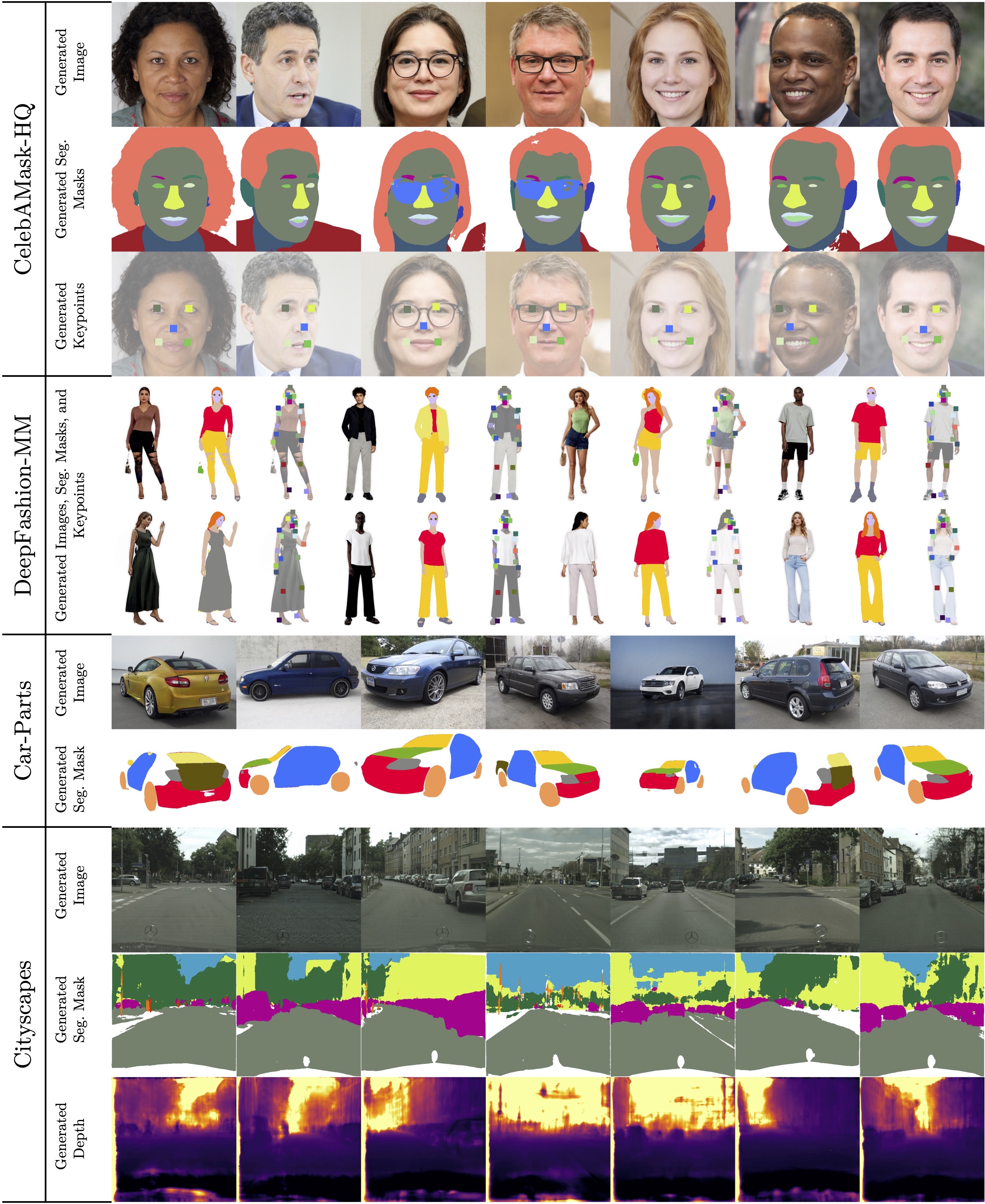}
    \caption{\small Examples of HandsOff generated labels (segmentation masks, keypoints, and depth) across four different domains. Generated labels capture fine details across various object orientations (CelebAMask-HQ, Car-Parts), object poses (DeepFashion-MM), and lighting conditions (Cityscapes). Note that HandsOff correctly assigns the label ``skin'' to the visible parts of the leg in the ripped areas of jeans (DeepFashion-MM, first row, first human) and correctly assigns the labels ``jacket'' and ``shirt'', despite the fact that the jacket and shirt are almost indistinguishable color-wise (DeepFashion-MM, first row, second human). Furthermore, generated keypoints are accurate despite partial occlusion, such as eyes behind glasses (CelebAMask-HQ, third and fourth image) or feet covered by long pants (DeepFashion-MM, second row, last human). HandsOff is also capable of identifying spatially small objects, such as street signs (Cityscapes, first, third, and fourth image).}
    \label{fig:generated_labels}
    \vspace{-4mm}
\end{figure*}

The HandsOff framework, shown in Fig.~\ref{fig:framework}, consists of three main components: (1) a generator (realized as a GAN), which maps a latent code $w \in \mathcal{W}$ to an image $X$, (2) an inverter, which maps an image $X$ to a latent code $w$, and (3) a label generator, which maps a latent code $w$ to a pixel-wise \textit{label} $Y$, such as a semantic segmentation mask.
HandsOff exploits the fact that the generator's latent space forms a rich, disentangled representation of images.
Since these latent spaces already encode semantically meaningful concepts from images \cite{shen2020interpreting, abdal2019image2stylegan, abdal2020image2stylegan++}, we aim to train a `label generator' that maps latents in this space to \textit{labels}.

Unfortunately, training this label generator requires paired data of latents $w$ with labels $Y$.
One approach, espoused by prior work~\cite{zhang21datasetgan}, could be to map the latent $w$ to an image $X$, and ask annotators to manually label the image.
However, in many applications, paired data of $(X, Y)$ is readily available, thanks to the careful efforts of dataset collectors.
Our key insight is that \textit{existing labeled image datasets can be used to train a label generator on GAN latent spaces,} using techniques from the GAN inversion literature. Below, we describe our specific approach for GAN inversion (Sec.~\ref{sec:framework:gan_inversion}), our representation of the GAN's latent space (Sec.~\ref{sec:framework:image_rep}), and finally, our label generator (Sec.~\ref{sec:framework:label_gen}).

\subsection{GAN inversion}\label{sec:framework:gan_inversion}
The key step in the HandsOff framework is to connect advances in GAN inversion to dataset generation.
GAN inversion allows us to use a small number of pre-existing labeled images to create a dataset of labeled \textit{latents}.
Our use of pre-existing labels allows practitioners to re-purpose existing labeled datasets, 
avoiding the cost of acquiring labels, including the prerequisite of maintaining annotation workstreams in their machine learning pipelines. 

Our GAN inversion is inspired by popular approaches in the image-editing community~\cite{ling2021editgan, zhu2016generative}.
Given a pre-trained generator $G$, we first train an encoder to predict a latent $w^{(e)}$ from an input image $X$.
In practice, this feed-forward encoder results in a good initial inversion of an image to a latent input.
To refine this initial estimate further, we solve the following regularized optimization problem:
\begin{align*}\label{eq:gan_inversion_opt}
    \min_{w : \|w - w^{(e)}\|_2^2 \leq c_{reg}} \mathcal{L}_{LPIPS}(X, G(w)) +
    &\lambda_{\ell_2}\|X - G(w)\|_2^2 
\end{align*}
where $\mathcal{L}_{LPIPS}$ is the Learned Perceptual Image Patch Similarity (LPIPS) loss \cite{zhang2018unreasonable}.
Although this problem is highly non-convex, in practice we find that using a fixed number of gradient descent iterations significantly refines the latent code.
This refinement step requires additional inference time, but this additional cost is incurred only once on a small number of training images.
In our experiments, we utilize ReStyle \cite{alaluf2021restyle} as the encoder, but we emphasize that our framework is amenable to \textit{any} GAN inversion procedure that does not modify the generator weights. Note that common approaches for GAN inversion fine-tune the \textit{generator} in order to achieve a better \textit{inversion} for a specific image~\cite{roich2022pivotal, alaluf2022hyperstyle, abdal2020image2stylegan++}.
To ensure our generator can produce \textit{new} images from the task domain, we keep the generator parameters frozen throughout the inversion process.

\subsection{Hypercolumn representation}\label{sec:framework:image_rep}
GAN inversion allows us to map images $X$ to latent codes $w$. 
We could use these latent codes directly to train a label generator that maps latent codes $w$ to labels $Y$.
However, this discards the rich representations encoded by the intermediate layers within the generator.
Rather than training on $w$ directly, we construct a hypercolumn representation $S^\uparrow$ from the generator's intermediate layers.
Specifically, we use a StyleGAN2 generator, where the latent code $w$ is used to modulate convolution weights in intermediate style blocks, which progressively grow an input to the final output image.
For a 1024 $\times$ 1024 resolution image, there are $L = 18$ style blocks. We utilize the approach of \cite{zhang21datasetgan} and take the intermediate output of these style blocks, upsample them channel-wise to the resolution of the full image, then concatentate each upsampled intermediate output channel-wise to obtain pixel-wise hypercolumns. 
Our final hypercolumn representation is denoted by $S^\uparrow$, with each pixel $j$ now having a hypercolumn $S^{\uparrow}[j]$ of dimension $C$. Due to the high dimensionality of the hypercolumns ($C = 6080$ for $1024 \times 1024$ images), we cap the generated image resolution to 512 $\times$ 512, and downsample intermediate outputs from higher resolutions. 

\subsection{Label generator}\label{sec:framework:label_gen}
The label generator exploits the semantically rich latent space of the generator to efficiently produce high quality labels for generated images. Because the latent codes already map to semantically meaningful parts of generated images, simple, efficient models suffice for generating labels.
Specifically, like in \cite{zhang21datasetgan}, we utilize an ensemble of $M$ multilayer perceptrons (MLPs).
The MLPs operate on a pixel-level, mapping a pixel's hypercolumn to a label.
To generate a label for a synthetic image, we pass the hypercolumn formed by latent code $w$ through the $M$ MLPs, and aggregate the outputs (via majority vote or averaging) to produce  a label. 
The $M$ MLPs are trained using a small number ($\sim$50) of pre-existing labeled images with a cross-entropy loss for generating discrete labels (e.g., segmentation masks) and mean-squared error loss for generating continuous labels (e.g., keypoint heatmaps).

Our use of an ensemble of MLPs naturally provides a way to filter out potentially poor labels by using the prediction uncertainty as a proxy for label quality. For discrete labels, we can utilize Jensen-Shannon divergence \cite{melville2005active, beluch2018power, kuo2018cost, zhang21datasetgan} across the $M$ MLPs to produce pixel-wise uncertainty maps. For predicting continuous labels, we compute the pixel-wise variance across the MLP outputs. In both cases, the overall image uncertainty is computed by summing across all pixels.

\section{Experimental results}\label{sec:exp}
\begin{table*}[t]
    \centering
    \label{tab:results_seg}
    \begin{tabular}{l c c c c c c}
        \toprule
        & \begin{tabular}{@{}c@{}} \# labeled \\ images\end{tabular} & \begin{tabular}{@{}c@{}}CelebAMask-HQ \\ 8 classes\end{tabular} & \begin{tabular}{@{}c@{}}Car-Parts \\ 10 train\end{tabular} & 
        \begin{tabular}{@{}c@{}}DeepFashion-MM \\ 8 classes\end{tabular} & 
        \begin{tabular}{@{}c@{}}DeepFashion-MM \\ 10 classes \end{tabular} &
        \begin{tabular}{@{}c@{}}Cityscapes \\ 8 classes \end{tabular}
        \\
        \midrule
        DatasetGAN & 16 & 0.7013 & $ \times $ & $ \times $ & $ \times $ & $ \times $ \\
        EditGAN  & 16 & 0.7244 &  0.6023 &  $ \times $ & $ \times $ & $ \times $\\
        Transfer Learning & 16 & 0.4575 & 0.3232 & 0.5192 & 0.4564 & 0.4954\\
        HandsOff (Ours) & 16 & \textbf{0.7814} & \textbf{0.6222} & \textbf{0.6094} & \textbf{0.4989} & \textbf{0.5510}\\
        \midrule 
        Transfer Learning & 50 & 0.6197 & 0.4802 & 0.6213 & 0.5559 & 0.5745\\
        HandsOff (Ours) & 50 & \textbf{0.7859} & \textbf{0.6679} & \textbf{0.6840} & \textbf{0.5565} & \textbf{0.6047}\\
        \bottomrule
    \end{tabular}
    \caption{\small{Downstream task performance for semantic segmentation tasks across various domains, reported in mIOU ($\uparrow$). HandsOff outperforms all baselines across all domains with both 16 and 50 labeled training images. $\times$ indicates a method that could not be run for a particular domain due to methodological shortcomings, such as requiring additional hand-labeled data.}\label{tab:segmentation_results}} 
\end{table*}
    
\begin{table*}
    \centering
    \resizebox{\linewidth}{!}{
    \begin{tabular}{l c c c c c c c c c c}
        \toprule
         &  \# labeled
            & \multicolumn{3}{c}{CelebAMask-HQ} 
            & \multicolumn{3}{c}{DeepFashion-MM}
            & \multicolumn{3}{c}{Cityscapes-Depth} \\
        & images & PCK-0.1 $\uparrow$ & PCK-0.05 $\uparrow$ & PCK-0.02 $\uparrow$ & PCK-0.1 $\uparrow$ & PCK-0.05 $\uparrow$ & PCK-0.02 $\uparrow$ & mNMSE $\downarrow$ & RMSE $\downarrow$ & RMSE-log $\downarrow$\\
        \midrule
        Transfer Learning & 16 & 78.96 & 42.06 & \phantom{0}7.32 & 91.24 & 83.52 & 48.21 & 0.4022 & 18.12 & 2.75 \\
        HandsOff (Ours) & 16 & \textbf{97.19} & \textbf{76.36} & \textbf{17.44} & \textbf{94.19} & \textbf{88.48} & \textbf{70.22} & \textbf{0.2553} & \textbf{14.52} & \textbf{1.64}\\ 
        \midrule 
        Transfer Learning & 50 & 90.88 & 61.75 & 12.30 & 91.24 & 83.52 & 48.20 & 0.2525 & 15.07 & 3.01 \\ 
        HandsOff (Ours) & 50 & \textbf{97.71} & \textbf{79.99} & \textbf{19.10} & \textbf{95.41} & \textbf{90.89} & \textbf{74.02} & \textbf{0.1967} & \textbf{13.01} & \textbf{1.58} \\
        \bottomrule
    \end{tabular}}
    \caption{\small{Downstream task performance for keypoint detection and depth estimation. HandsOff outperforms all other methods when trained on 16 or 50 labeled images, demonstrating an impressive ability in generating \textit{continuous}-valued keypoint heatmaps and depth maps.} \label{tab:results_kp_depth}} 
\end{table*}

We extensively evaluate HandsOff in generating both discrete (segmentation masks) and continuous (keypoint heatmaps and depth) labels across four challenging domains: Faces, Cars, Full-Body Human Poses, and Urban Driving Scenes.
We utilize various pre-trained StyleGAN2 generators \cite{karras2020analyzing, fu2022styleganhuman, Gadde2021} and ReStyle inverters \cite{alaluf2021restyle}. To train the label generator, we utilize existing labels from CelebAMask-HQ \cite{CelebAMask-HQ}, Car-Parts \cite{DSMLR_Carparts}, DeepFashion-MultiModal \cite{jiang2022text2human, liuLQWTcvpr16DeepFashion}, and Cityscapes \cite{Cordts2016Cityscapes}. The key assumption of HandsOff is that GAN inverted image reconstructions align well with the original labels. We present visualizations of reconstructed image alignment in Appendix~\ref{sec:add_results:recon_align}.
Our label generator architecture across all domains and tasks is an $M = 10$ ensemble of 2-hidden layer MLPs. This simple architecture is a distinct strength of the HandsOff framework: intensive parameter and architecture finetuning are not necessary to achieve state-of-the-art empirical performance. For the label generator, we provide training details in Appendix~\ref{sec:exp_setup:training_setup}, architecture details in Appendix~\ref{sec:exp_setup:label_generator}, and ablations in Appendix~\ref{sec:ablations}.

\subsection{Experimental set-up}\label{sec:exp:setup}
\paragraph{Downstream network} In all domains and tasks, we utilize DeepLabV3 with a ResNet151 backbone as our downstream network. We generate 10,000 synthetic images and labels, filter out the top 10\% most uncertain images (see Sec.~\ref{sec:framework:label_gen}), and train our downstream network for 20 epochs with the 9,000 remaining images. For segmentation, we have DeepLabV3 output a probability distribution over all of the parts for each pixel, whereas for keypoints or depth, we have DeepLabV3 output continuous values. Due to the dynamic nature of elements in the Cityscapes dataset, slight imperfections in the reconstructions uniquely affect segmentation mask alignment. To mitigate this, we perform an extra fine tuning step with the original 16 or 50 labeled examples used to train the label generator while training for semantic segmentation. Training details for the downstream network can be found in Appendix~\ref{sec:exp_setup:training_setup} and ablations can be found in Appendix~\ref{sec:ablations}.

\paragraph{Baselines} We compare HandsOff against three baselines: DatasetGAN, EditGAN, and Transfer Learning. We are only able to evaluate DatasetGAN in the face domain, as DatasetGAN is unable to accommodate the change in labeling scheme from their custom labeled car dataset to the larger Car-Parts-Segmentation dataset, thus highlighting another drawback of requiring GAN labeled images. For EditGAN, we adopt the image editing framework to synthesize labels for images. However, we are unable to test in the full-body human poses and urban driving scene domains, as EditGAN has only released checkpoints for the face and car domains. For the Transfer Learning baseline, we initialize DeepLabV3 with pretrained weights on ImageNet, then finetune the classification head of the model on the 16 or 50 labeled images used to train HandsOff until convergence. This baseline is used to benchmark our method, which is trained on up to 50 labeled images, against a model that is trained on 100,000+ \textit{labeled} out-of-domain images in addition to the 16 or 50 labeled in-domain images. 

\paragraph{Datasets} For faces, we split CelebAMask-HQ into a set of 50 training, 450 validation, and 29,500 testing images. We collapse the 19 original segmentation classes into 8 and scale the keypoint locations in the low resolution version of images found in CelebA to the full resolution images. For cars, we retain the original 400 image train set, split the test set into a set of 20 images for validation and 80 images for testing, and collapse the 19 original classes into 10. For full-body human poses, we split DeepFashion-MultiModal into a set of 200 training, 500 validation, and 12,000 testing images. We collapse the 24 original segmentation classes into 8 and 10 classes and retain the original 21 labeled keypoint locations. For Cityscapes, because the ground truth test labels are not released, we split 300 and 1275 images from the original train set for validation and test, respectively. We utilize the eight groups (e.g., human, vehicle, etc) as our class labels. Note that while our train sets may contain more than 50 images, we use \textit{at most} 50 labeled images from the train sets to train HandsOff in each domain. Details on class collapse can be found in Appendix~\ref{sec:exp_setup:seg_mask_collapse}.

\subsection{HandsOff generated datasets}\label{sec:exp:vis} We visualize the generated image-label pairs from HandsOff in Fig.~\ref{fig:generated_labels}. HandsOff is capable of generating very high quality labels across all domains. In the face domain, HandsOff is capable of producing segmentation masks that can correctly distinguish left/right features like eyes or ears and identify rare occurring classes such as glasses. Furthermore, it produces extremely accurate keypoint locations even when such locations may be partially occluded. Within the full-body human pose domain, HandsOff produces finely detailed segmentation masks, best illustrated by the segmentation mask for the first human in the top row of Fig.~\ref{fig:generated_labels}, who is wearing a pair of ripped jeans and the second human in the top row who is wearing the same colored jacket and shirt (see caption for more details).  Generated labels are consistently high quality across a diverse array of object orientations, as seen in the various face rotations, human poses, or car orientations of Fig.~\ref{fig:generated_labels}. Finally, in extremely complex scenes, such as Cityscapes, HandsOff produces labels for visually minuscule classes, such as street lamps or traffic signs. Additional examples of generated labels can be found in Appendix~\ref{sec:add_results:label_examples}.

\begin{figure*}[t]
\centering
\subcaptionbox{\label{fig:car_longtail_sub}}[0.31\textwidth]{\includegraphics[width=\linewidth]{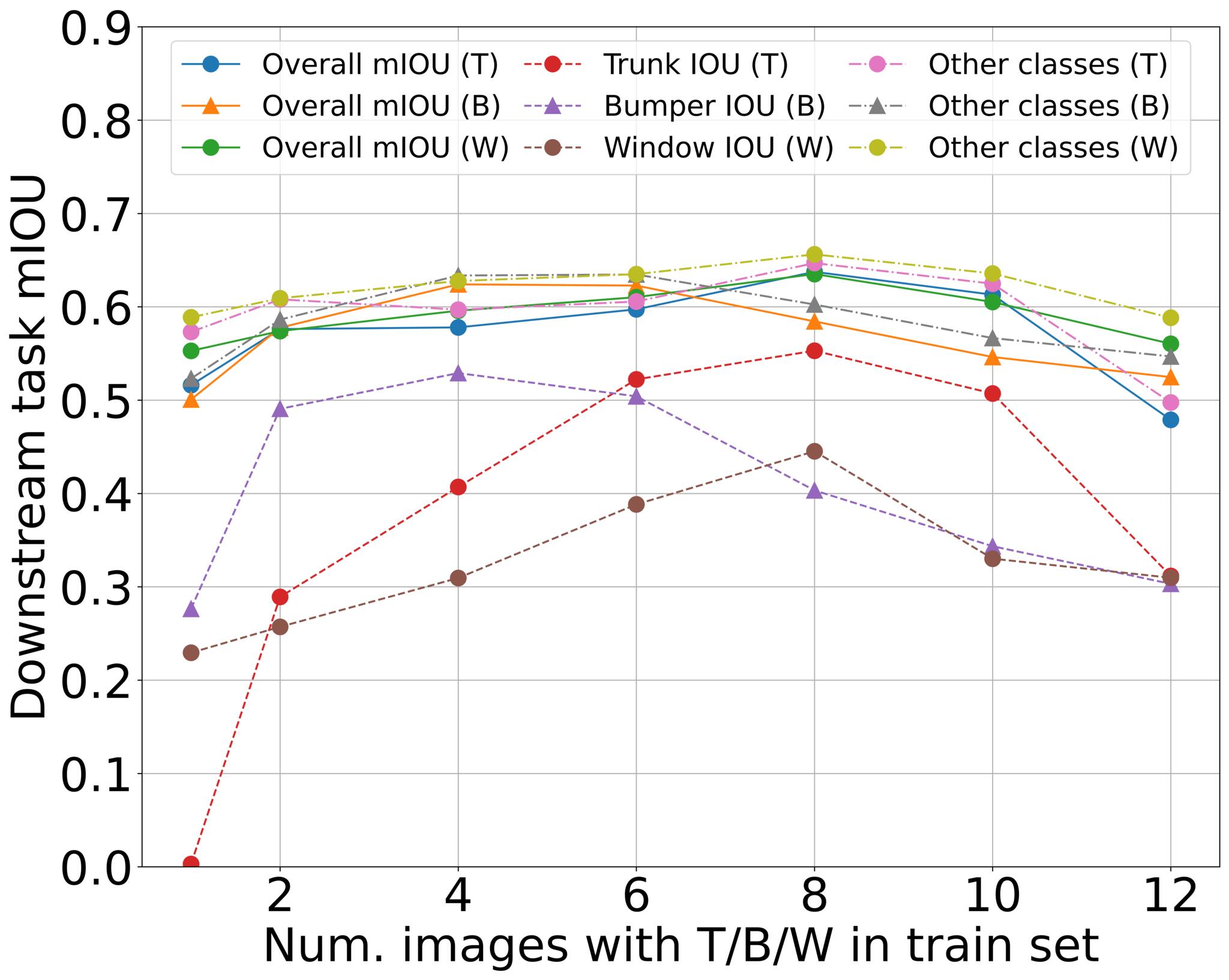}}
\hfill
\subcaptionbox{\label{fig:face_longtail_sub}}[0.31\textwidth]
{\includegraphics[width=\linewidth]{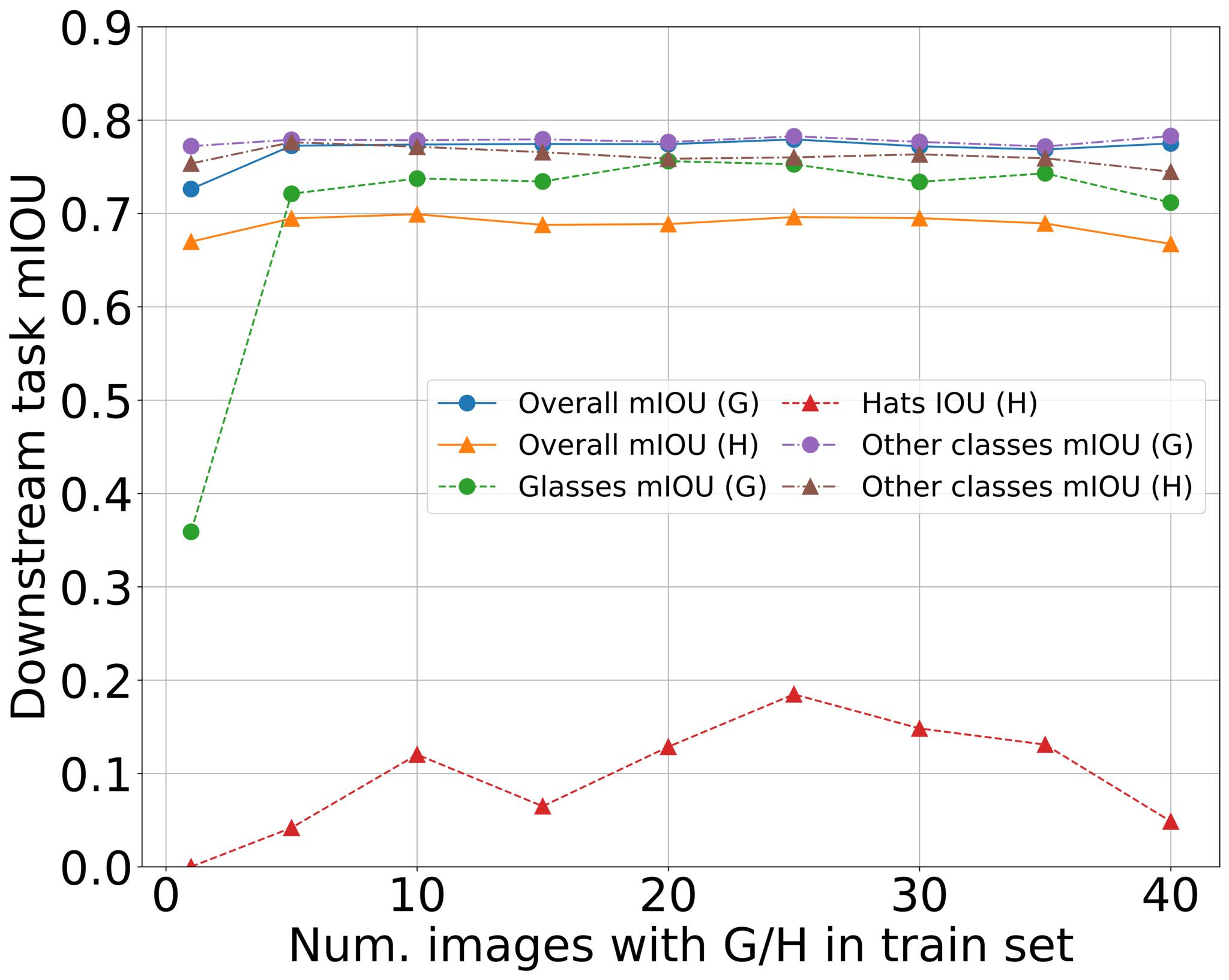}}
\hfill
\subcaptionbox{\label{fig:face_longtail_add}}[0.31\textwidth]
{\includegraphics[width=\linewidth]{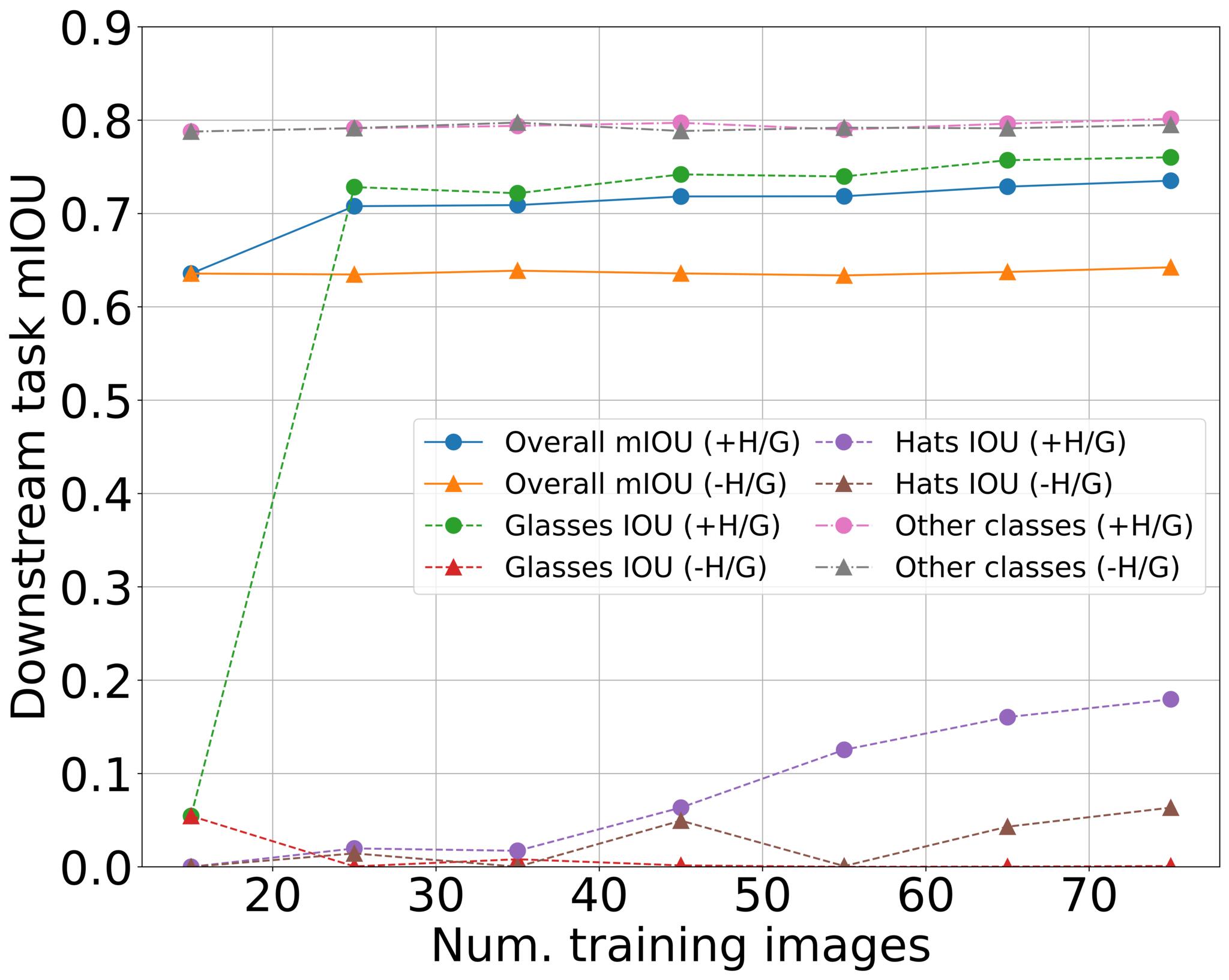}}

\caption{\small{Substitution experiments for various long-tail parts; (a) in cars  - trunk (T), back bumper (B), back window (W); (b) in faces - glasses (G), hats (H). As the proportion of images containing the long-tail part increases in the training set, the performance of the long-tail class improves until it enters the \textit{overfitting} regime. Non-long-tail mIOU tracks closely with overall IOU, implying dramatic gains in long-tail IOU do not come at the expense of other parts. (c) Addition experiments for face long-tail parts. (+H/G) indicates that images containing hats and images containing glasses are added to a base set, while (-H/G) indicates images containing neither hats nor glasses are added. The long-tail IOU of both parts \textit{simultaneously} increase as images containing hats and images containing glasses are added to the base training set, with no negative impact on the performance of other classes.}}
\end{figure*}

\subsection{Segmentation results}\label{sec:exp:segmentation}
As seen in Tab.~\ref{tab:segmentation_results}, we achieve \textbf{state-of-the-art performance} on synthetic data trained semantic segmentation in all four domains, as measured in mean Intersection-over-Union (mIOU). Specifically, HandsOff outperforms DatasetGAN by 11.4\% and EditGAN by 7.9\% in the face domain when trained with \textit{the same number of labeled images}. Increasing the number of labeled training images for HandsOff results in further performance gains, with 12.1\% and 8.5\% improvements over DatasetGAN and EditGAN, respectively. Unlike DatasetGAN, we are able to increase the number of labeled training images without incurring the associated costs of collecting new human annotated images. We emphasize again that with new domains, such as full-body human poses or urban driving scenes, it is not possible to train DatasetGAN-based frameworks as they rely on manual labels for GAN generated images. 
Therefore, we benchmark against the transfer learning baseline in these domains.
Notably, HandsOff outperforms the transfer learning baseline by 17.4\% (full-body human poses) and 11.2\% (urban driving scenes) when both methods are trained on 16 labeled images; and 10.1\% (full-body human poses) and 5.3\% (urban driving scenes) when trained on 50 images.

\subsection{Keypoint and depth results}\label{sec:exp:kp_depth}
We utilize HandsOff to generate \textit{continuous} valued labels for keypoints and depth tasks. As seen in Tab.~\ref{tab:results_kp_depth}, we demonstrate strong empirical performance in generating both keypoints and depth maps. To synthesize keypoints, we utilize the keypoint heatmap regression frramework, where our label generator is asked to output a continuous-valued spatial heatmap for each keypoint. See Appendix~\ref{sec:exp_setup:keypoint} for a detailed explanation of keypoint regression. For downstream task performance, we report the Percentage of Correct Keypoints (PCK) for different threshold values $\alpha$, denoted PCK-$\alpha$. For a keypoint to be predicted correctly, the estimate must be no further from the true keypoint than $\alpha \cdot \max\{h, w\}$, where $h$ and $w$ are the height and width of the minimum size bounding box that contains all of the keypoints. We note that even for small $\alpha$ (i.e., $\alpha = 0.02)$, HandsOff is able to correctly predict 2.4$\times$ and $1.5\times$ more keypoints than the transfer learning baseline in the face and full-body human pose domains, respectively. This implies that HandsOff is able to predict keypoints up to an extremely tight radius of the original keypoint location compared to other methods. 

For depth, we report masked normalized mean-squared error (mNMSE), root mean-squared error (RMSE), and root mean-squared error of the log-depth values (RMSE-log). Because Cityscapes depth maps contain corrupted depth values, we train HandsOff only non-corrupted pixels. Furthermore, to compute mNMSE, we compute the normalized mean-squared error only on the non-corrupted pixels. That is, let $\widehat{\vy}$ and $\vy$ are the predicted and true depth maps, respectively, and $\mM$ be a mask indicating the non-corrupted elements of $\vy$. mNMSE is computed as $\frac{\norm{\widehat{\vy}_{\mM} - \vy_{\mM}}_2^2}{\norm{\vy_{\mM}}_2^2}$, where $\va_{\mM}$ denotes the depth map $\va$ at non-corrupted locations. When reporting RMSE and RMSE-log, we adopt the standard practice \cite{casser2019unsupervised, watson2021temporal} in depth estimation of cropping the middle 50\% of the image and clamping predicted depth values to be within $0.001$ and $80$ before computing RMSE and RMSE-log values. As shown in Tab.~\ref{tab:results_kp_depth}, HandsOff is able to achieve a sizable advantage in all three metrics, outperforming transfer learning, resulting in 36.5\%, 19.9\%, and 40.27\% decreases in mNMSE, RMSE, and RMSE-log when trained on 16 labeled images and 22.1\%, 13.6\%, and 47.6\% decreases when trained on 50 labeled images.

\begin{figure*}[t]
    \centering
    \includegraphics[width=0.8\textwidth]{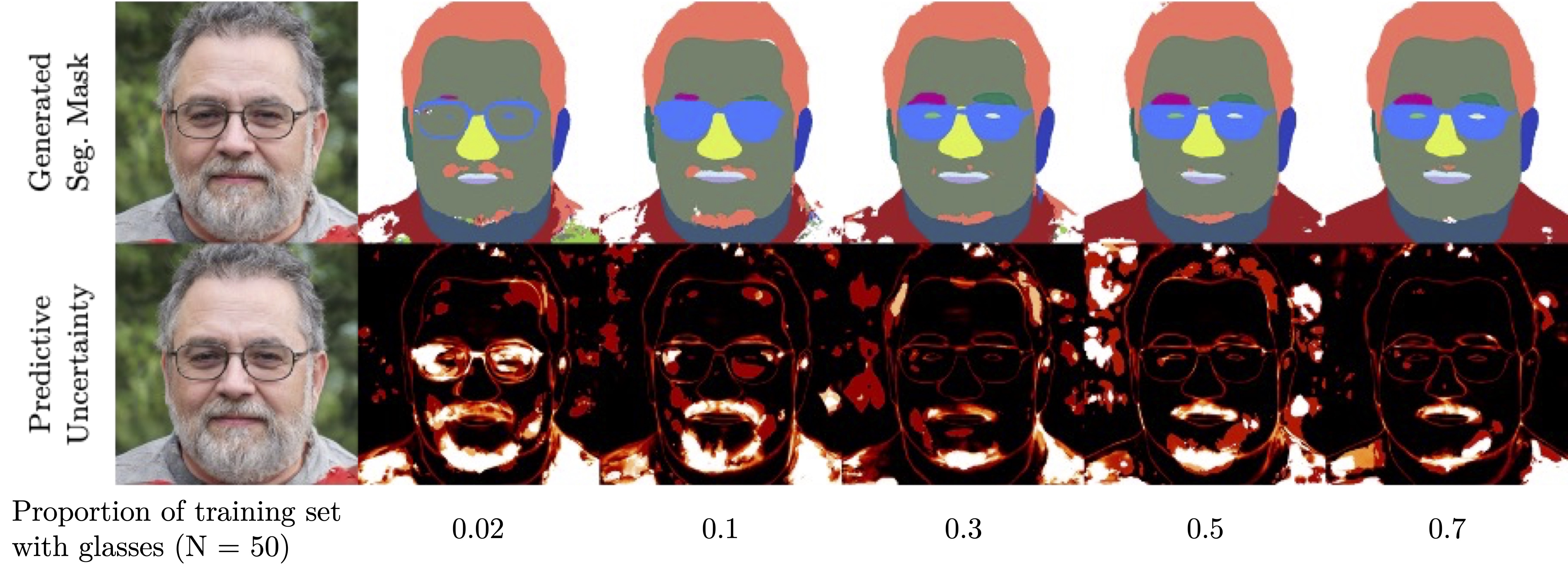}
    \caption{\small{Visualization of generated segmentation mask (top row) and pixel-wise label generator uncertainty (bottom row) as the proportion of the training set containing the glasses increases. Not only do we see qualitative improvement in the generated label for glasses, we also see that the classifier is less \textit{uncertain} when generating the correct label.}}
    \label{fig:longtail_vis}
\end{figure*}

\subsection{Long-tail semantic segmentation}\label{sec:exp:long_tail}
The HandsOff framework's ability to generate high quality synthetic datasets unlocks new degrees of freedom for model development previously unachievable with fixed, hand-annotated datasets. We now explore one example: mitigating the effects of the long-tail common in semantic segmentation datasets. For CelebAMask-HQ, images with hats and glasses make up less than 5\% of the 30,000 labeled images, and a similar situation exists with trunks, back bumpers, and back windows in the Car-Parts dataset. These examples form the long-tail classes of their respective datasets, and their rare occurrence during training results in poor model performance at evaluation time. 

The HandsOff framework altogether sidesteps this limitation of traditional datasets: by generating labeled synthetic images, we can control the occurrence of rare classes in our training data and significantly mitigate the effects of the long-tail. Because training the label generator requires less than 50 annotated images, we only require 5-10 occurrences of long-tail classes in order to generate an unlimited number of those occurrences in our synthetic dataset.  Our experiments precisely quantify the small number of annotated examples of rare classes required to significantly improve downstream task performance on those classes. They fall into two categories: \textbf{Substitution} experiments, that fix a total number of training images and vary the proportion of rare class occurrences, and \textbf{Addition} experiments, that grow the size of the training set by adding images with rare classes. The substitution experiments ensure that any gains in the performance of identifying the long-tail class are not a by-product of increasing training set size. We perform substitution experiments considering only one long-tail part at a time. On the other hand, the addition setting is indicative of how a practitioner would deploy HandsOff: starting with a base set of labeled training images and further augmenting it with images containing rare classes deemed crucial to identify. To mirror what often happens in practice, we perform addition experiments by adding images containing multiple long-tail classes at a time. 

\textbf{Substitution.} We begin with an initial set of 16 (cars) or 50 (faces) labeled images containing one image of the rare part, and then vary the proportion of the rare part. As seen in Fig.~\ref{fig:car_longtail_sub} and~\ref{fig:face_longtail_sub}, a small proportion of rare classes results in poor class identification performance, but as the proportion of images with long-tail classes increases, the long-tail part IOU increases by as much as 0.55 for car trunks and 0.40 for face glasses before eventually plateauing. We note that hats are a particularly challenging part to generate labels for due to the diversity of their size, shape, color, and orientation. Nevertheless, we still see a sizable increase of 0.2 IOU. We additionally plot the overall mIOU and the mIOU of non-long-tail parts to demonstrate that modifying the composition of the training set does not hurt performance on non-long-tail parts. In other words, shifting the training set part distribution to an extent has negligible impacts on the performance of non-long-tail parts, while resulting in large gains in long-tail class detection. Beyond proportions of $\sim$0.7, further increasing the proportion of the training set eventually causes drops in both long-tail part IOU and the mIOU of non-long-tail parts, owing to the label generator hallucinating long-tail classes where they do not belong. The impacts of substituting images with long-tail classes are best illustrated in Fig.~\ref{fig:longtail_vis}. As the proportion of images with glasses grows, the generated mask captures glasses with increasing accuracy, eventually even distinguishing eyes that are visible through the glasses. Underneath the segmentation masks, we showcase the pixel-wise label generator uncertainty measured by Jensen-Shannon divergence (See Sec.~\ref{sec:framework:label_gen}). Not only does the generated label improve qualitatively, the label generator is \textit{less uncertain} about the region of the image corresponding to the glasses. Additional visual examples of both segmentation mask and label generator uncertainty can be found in Appendix~\ref{sec:add_results:longtail}.

\textbf{Addition.} We augment a small training set of 15 images with additional images containing hats or glasses. Fig.~\ref{fig:face_longtail_add} demonstrates significant IOU increases (+0.71) in long-tail classes. The figure further highlights that these increases are not simply due to additional examples: targeted additions outperform the scenario where we add the same number of images, but the added images do \textbf{not} contain hats or glasses.
These improvements in long-tail classes do not come at the expense of performance in other classes, as demonstrated by the overall mIOU and mIOU of non-long-tail classes. Unlike the substitution experiments, these performance improvements do not eventually drop, since the number of training examples continues to increase.

Our experiments showcase the power of the HandsOff framework to mitigate the long-tail problem. By explicitly including images with the long-tail class in our label generator training data, we are able to bridge the gap between performance in rare and common classes. The number of images with long-tail classes necessary to generate high quality labels of the long-tail is even smaller than the already small number of images needed to train HandsOff, meaning that the gains in long-tail class performance essentially come for free. If the long-tail class has been deemed crucial to identify, then it is likely that a practitioner has access to $\sim$20 labeled images containing the long-tail class. The performance gains in long-tail performance achieved by HandsOff are not practically replicable in DatasetGAN, where human supervision is needed to both identify generated images containing the long-tail class and provide precise pixel-level annotations.

\section{Discussion}\label{sec:discussion}
We present the HandsOff framework, which produces high quality labeled synthetic datasets without requiring further annotation of images for a multitude of tasks across various challenging domains. 
HandsOff achieves state-of-the-art performance over several recent baselines when training a downstream network with our synthetically generated data.
Furthermore, HandsOff enables user control of the training data composition, leading to dramatic performance gains in long-tail semantic segmentation.
This suggests that HandsOff can play a vital role in curtailing the effects of the long-tail.
While synthetic datasets have the potential to supplant human annotations, they can also complement them. We leave as future work to investigate the collaborative power of having a human-in-the-loop refine synthetically generated annotations, and bring about the best of both worlds.

\clearpage
\newpage
{\small
\bibliographystyle{ieee_fullname}
\bibliography{bib}
}

\clearpage
\appendix
\section{Acknowledgements}\label{sec:ack}
We thank Anurag Beniwal, Benjamin Biggs, Kevin Chen, Mark Davenport, Peter Hallinan, Gerard Medioni, Vaishaal Shankar, Scott Sun, and Guanglei Xiong for their helpful comments, pointers, and feedback.

\section{Experimental setup}\label{sec:exp_setup}
\subsection{Dataset details}\label{sec:exp_setup:dataset_details}
In our experiments, we utilize the following datasets. We report the licenses for all datasets that publicly list them.
\begin{itemize}
    \item CelebAMask-HQ \cite{CelebAMask-HQ}. License: non-commercial research and educational purposes.
    \item Car-Parts \cite{DSMLR_Carparts}.
    \item DeepFashion-MultiModal \cite{liuLQWTcvpr16DeepFashion, jiang2022text2human}. License: non-commercial research purposes.
    \item SHHQ \cite{fu2022styleganhuman}. License: CC0 and free for research use.
    \item Cityscapes \cite{Cordts2016Cityscapes}. License: on-commercial research and educational purposes.
\end{itemize}
We also utilize pre-trained StyleGAN2 and ReStyle models. In the face and car domain, these models were trained on the following datasets:
\begin{itemize}
    \item FHHQ \cite{karras2019style}. License: Creative Commons BY-NC-SA 4.0 license by NVIDIA Corporation.
    \item LSUN \cite{yu2015lsun}. 
    \item Stanford Cars \cite{krause2013cars}. License: non-commercial research and educational purposes.
\end{itemize}
To make the DeepFashion-MultiModal segmentation masks compatible with StyleGAN-Human, we first used the segmentation mask to determine the background for each image and set the background to white. We then re-sized each image to the same size SHHQ images.

\subsection{Segmentation mask class collapse}\label{sec:exp_setup:seg_mask_collapse}
Consistent with prior works \cite{zhang21datasetgan}, we collapse the original labels in each dataset into a smaller number of labeled parts. For CelebAMask-HQ dataset, we remove any distinction between left/right in a number of parts (e.g., ears, eyes, eyebrows). Furthermore, we form one mouth part consisting of upper/lower lips and mouth. Finally, we collapse all accessories and clothing into background. See Tab.~\ref{tab:class_collapse_faces} for exact class collapse mapping. In long-tail experiments, we un-collapse the relevant long-tail classes (glasses and hats) and consider them separate classes.

For the Car-Parts dataset, we remove any distinction between left/right and front/back for parts such as doors, lights, bumpers, and mirrors. We also merge trunks and tailgates to be the same class. See Tab.~\ref{tab:class_collapse_cars} for exact class collapse mapping.

For DeepFashion-MultiModal, we consider two degrees of class collapse. In the first, we consider the following ten classes, with original classes included in parentheses: tops (tops and ties), outerwear, dresses (dresses, skirts, rompers), bottoms (pants, leggings, belts), face (face, glasses, earrings), skin (skin, neckwear, rings, wrist accessories, gloves, necklaces), footwear (shoes and socks), bags, and hair (hair and headwear). In the second, we further collapse the classes by including outerwear in tops and bags as background. See Tab.~\ref{tab:class_collapse_fashion8} and ~\ref{tab:class_collapse_fashion10} for exact class collapse mappings.

For Cityscapes, we utilize the eight groups listed on the Cityscapes official website as our classes, with slight modifications. We consider parts labeled sidewalk, parking, and rail track as a part of the void class. See Tab.~\ref{tab:class_collapse_city} for exact class collapse mapping.

\begin{table*}
    \begin{subtable}[t]{0.48\linewidth}
        \begin{tabular}[t]{l l}
             \begin{tabular}{@{}l@{}}Collapsed\\ 
             label (8) \end{tabular} & CelebAMask-HQ original labels  \\
             \toprule
             Background & \begin{tabular}{@{}l@{}}Background (0), hat (14), earring (15),\\ 
             necklace (16), neck (17), clothes (18)\end{tabular} \\
             Skin & Skin (1)\\
             Nose & Nose (2)\\
             Eyes & Left eye (3), right eye (4), glasses (5)\\
             Eyebrows & Left eyebrow (6), right eyebrow (7) \\
             Ears & Left ear (8), right ear (9)\\
             Mouth & Mouth (10), upper lip (11), lower lip (12)\\
             Hair & Hair (13)
        \end{tabular}
        \caption{}
        \label{tab:class_collapse_faces}
    \end{subtable}
    \hspace{\fill}
    \begin{subtable}[t]{0.48\linewidth}
        \begin{tabular}[t]{l l}
             \begin{tabular}{@{}l@{}}Collapsed\\ 
             label (10) \end{tabular} & Car-Parts original labels  \\
             \toprule
             Background & Background(0)\\
             Bumper & Back bumper (1), front bumper (7)\\
             Back window & Back glass (3)\\
             Doors & \begin{tabular}{@{}l@{}}Back left door (3), back right door (5),\\ 
             front left door (9), front right door (11)\end{tabular} \\
              
             Lights & \begin{tabular}{@{}l@{}}Back left light (4), back right light (6),\\ 
             front left light (10), front right light (12)\end{tabular}\\
             Windshield & Front glass (8)\\
             Hood & Hood (13)\\
             Mirror & Left mirror (14), right mirror (15)\\
             Trunk & Tailgate (16), trunk (17)\\
             Wheel & Wheel (18) 
        \end{tabular}
        \caption{}
        \label{tab:class_collapse_cars}
    \end{subtable}

    \bigskip

    \begin{subtable}[t]{0.48\linewidth}
        \begin{tabular}[t]{l l}
             \begin{tabular}{@{}l@{}}Collapsed\\ 
             label (10) \end{tabular} & DeepFashion-MM original labels  \\
             \toprule
             Background & Background(0)\\
             Top & Top (1), tie (23)\\
             Outerwear & Outerwear (2)\\
             Dress & Skirt (3), dress (4), romper (21) \\
              
             Bottoms & Pants (5), leggings (6), belt (10)\\
             
             Face & Glasses (8), face (14), earring (22)\\
             Skin & \begin{tabular}{@{}l@{}}Neckwear (9), skin (15), ring (16),\\ 
             Wrist accessories (17), gloves (19), \\
             necklace (20)\end{tabular}\\
             
             Footwear & Footwear (11), socks (18)\\
             Bags & Bags (12)\\
             Hair & Headwear (7), hair (13) 
        \end{tabular}
        \caption{}
        \label{tab:class_collapse_fashion8}
    \end{subtable}
    \hspace{\fill}
    \begin{subtable}[t]{0.48\linewidth}
        \begin{tabular}[t]{l l}
             \begin{tabular}{@{}l@{}}Collapsed\\ 
             label (8) \end{tabular} & DeepFashion-MM original labels  \\
             \toprule
             Background & Background(0), bags(12)\\
             Top & Top (1), tie (23),  outerwear (2)\\
             Dress & Skirt (3), dress (4), romper (21) \\
              
             Bottoms & Pants (5), leggings (6), belt (10)\\
             
             Face & Glasses (8), face (14), earring (22)\\
             Skin & \begin{tabular}{@{}l@{}}Neckwear (9), skin (15), ring (16),\\ 
             Wrist accessories (17), gloves (19), \\
             necklace (20)\end{tabular}\\
             
             Footwear & Footwear (11), socks (18)\\
             Hair & Headwear (7), hair (13) 
        \end{tabular}
        \caption{}
        \label{tab:class_collapse_fashion10}
    \end{subtable}

    \bigskip
        
    \begin{subtable}[t]{0.96\linewidth}
        \begin{tabular}[t]{l l}
             Collapsed label (8) & Cityscapes (Fine annotations) original labels  \\
             \toprule
             Void & \begin{tabular}{@{}l@{}}Unlabeled (0), ego vehicle (1), rectification border (2), out of ROI (3), static (4), dynamic (5), \\ 
             ground (6), sidewalk (8), parking (9), rail track (10)\end{tabular}\\

             Road & Road (7) \\
             Construction & Building (11), wall (12), fence (13), guard rail (14), bridge (15), tunnel (16)\\
             Object & pole (17), polegroup (18), traffic light (19), traffic sign (20)\\
             Nature & Vegetation (21), terrain (22)\\
             Sky & Sky (23)\\
             Human & Person (24), rider (25)\\
             Vehicle & \begin{tabular}{@{}l@{}}UCar (26), truck (27), bus (28), caravan (29), trailer (30), train (31), motorcycle (32),\\ 
             bicycle (33), license plate (-1)\end{tabular}\\ \\
        \end{tabular}
        \caption{}
        \label{tab:class_collapse_city}
    \end{subtable}
    \caption{Mapping from collapsed class label to original class label in faces (a), cars (b), full-body human poses (c), (d), and urban driving scenes (e) domains. Original class numbers provided for each original class label name in parentheses.}
    \vspace{10mm} 
\end{table*}

\subsection{Training setup}\label{sec:exp_setup:training_setup}
All experiments were run on V100 GPUs using Amazon Web Services
(AWS) P3dn.24xlarge instances. Each MLP in the label generator ensemble was trained with the same parameters for all domains and tasks. Each MLP was trained for $\sim 4$ epochs via the Adam optimizer \cite{kingma2014adam} with learning rate 0.001 and batch size 64. For all results presented in Tab.~\ref{tab:segmentation_results} and~\ref{tab:results_kp_depth}, the labeled images used to train the label generator were chosen at random. For long-tail experiments (Sec.~\ref{sec:exp:long_tail}), images with the long-tail part were identified. Then, the labeled training images were selected at random from the identified images. 

Prior to training the downstream network, we filter out the top 10\% most uncertain synthetically generated images, except for the long-tail experiments. No filtering is performed for long-tail experiments to ensure that images with long-tail parts, which are more likely to be ``uncertain'', are included in the training set for the downstream network. To train the downstream network, we again utilize the Adam optimizer \cite{kingma2014adam} with learning rate 0.001 and batch size $64$. We train ReStyle \cite{alaluf2021restyle} on the set of labeled training images randomly selected from SHHQ \cite{liuLQWTcvpr16DeepFashion, hacheme21} and Cityscapes \cite{Cordts2016Cityscapes} for the full-body human poses and urban driving scene domains, respectively. We use default settings found in the ReStyle repository.

\subsection{GAN inversion setup}\label{sec:exp_setup:gan_inversion}
For the full-body human poses and urban driving scenes domains, we train ReStyle with the candidate training examples. Our framework only uses GAN inversion to obtain latent codes for training the label generator. Training on the candidate training examples thus ensures that ReStyle optimally reconstructs these latent codes. For faces and cars, this procedure is not necessary because ReStyle optimally reconstructs the latent codes of training examples without training. For the optimization-based finetuning, we utilize $c_{reg} = 0.5$ and $\lambda_{\ell_2} = 0.1$ for all domains. We run 300 optimization steps for the car domain, 500 iterations for the face and urban driving scenes domains, and 2,000 iterations for the human full-body poses domain. See Appendix~\ref{sec:ablations} for ablations on GAN inversion optimization steps.

\subsection{Label generator architecture}\label{sec:exp_setup:label_generator}
For all experiments, we utilize an ensemble of two layer MLPs with ReLU activations and batch normalizations for our label generator. We sweep the combination of layer widths and report the performance associated with the best performing combination for each domain and number of labeled training images. See Appendix~\ref{sec:ablations} for ablations on layer widths. Below, we report the combination of label generator sizes that produced the best performance. $(x, y)$ indicates that a network with first hidden layer of width $x$ and second hidden layer of width $y$ was used.

\paragraph{Faces} For segmentation, we utilize layer sizes of (256, 32) for 50 training images and (512, 64) for 16 training images. For keypoints, we utilize (512, 32) for PCK-0.1, PCK-0.05, and PCK-0.02 with 50 training images. For 16 training images, we utilize (512, 64) for PCK-0.1 and (512, 32) for PCK-0.05 and PCK-0.02. 

\paragraph{Cars} For segmentation, we utilize (512, 256) for both 50 training images and 16 training images. 

\paragraph{Full-body human poses} For segmentation, we utilize (1024, 32) and (2048, 64) for 50 training images in the 8 class and 10 class settings and (2048, 64) and (2048, 128) for 16 training images in the 8 class and 10 class settings. For keypoints, we utilize (512, 128), (256, 128), and (128, 64) for PCK-0.1, PCK-0.05, and PCK-0.02 with 50 training images. For 16 training images, we utilize (512, 256) for all three PCK thresholds.

\paragraph{Urban driving scenes} For segmentation, we utilize (512, 64) for both 50 and 16 training images. For depth maps, we utilize (512, 256) for both 50 and 16 training images.

\subsection{Keypoint heatmap regression}\label{sec:exp_setup:keypoint}
For keypoint detection experiments, we utilize a heatmap regression setup. Given an image (of size $H \times W)$ and a corresponding list of $K$ keypoints, we form a corresponding pixel-wise label for the image as follows. For each of the $K$ keypoints, we create a $H \times W$ sized heatmap. The values of the heatmap are the values of the density of a standard two-dimensional Gaussian centered at the location of the keypoint with variance $\sigma$. We further scale the values of the heatmap by 10, so that the maximum value of the heatmap is $10$. We find through hyperamater tuning that $\sigma = 25$ works well for full body while $\sigma=5$ works well for faces. With faces, we use $\sigma=5$ for the original sized CelebA images and then resize the mask to be of CelebAMask-HQ resolution. 

The label generator and downstream task are tasked with predicting a vector of $K$ values for each pixel. At test time, after predicting $K$ heatmaps corresponding to the $K$ keypoints, we take the location of the maximum element of each heatmap as the location of the keypoint. When computing the PCK metric, we only compute if a keypoint was correctly detected for visible keypoints. Information on if a particular keypoint is visible or not is provided in DeepFashion-MM, but not for CelebA.

\section{Ablation studies}\label{sec:ablations}
In this section, we present ablation studies that shed insights on various hyperparameters. 

\begin{figure*}[t]
\centering
\hfill
\subcaptionbox{\small{Ablation for hypercolumn dimension in the face domain.}\label{fig:hypercolumn_dim}}[0.31\textwidth]{\includegraphics[width=\linewidth]{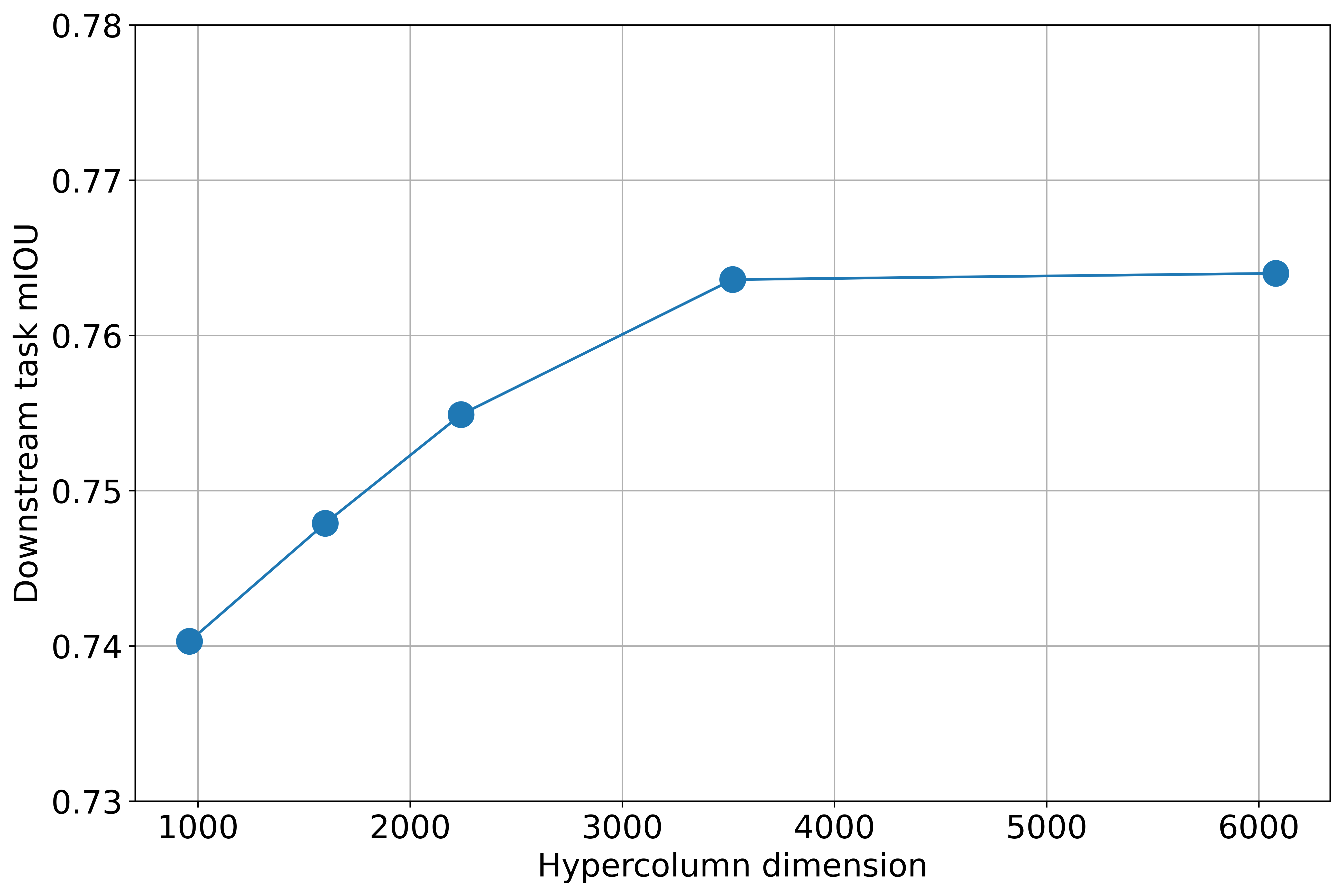}}
\hfill
\subcaptionbox{\small{Ablation for ensemble size in the face domain.}\label{fig:ensemble_size}}[0.31\textwidth]
{\includegraphics[width=\linewidth]{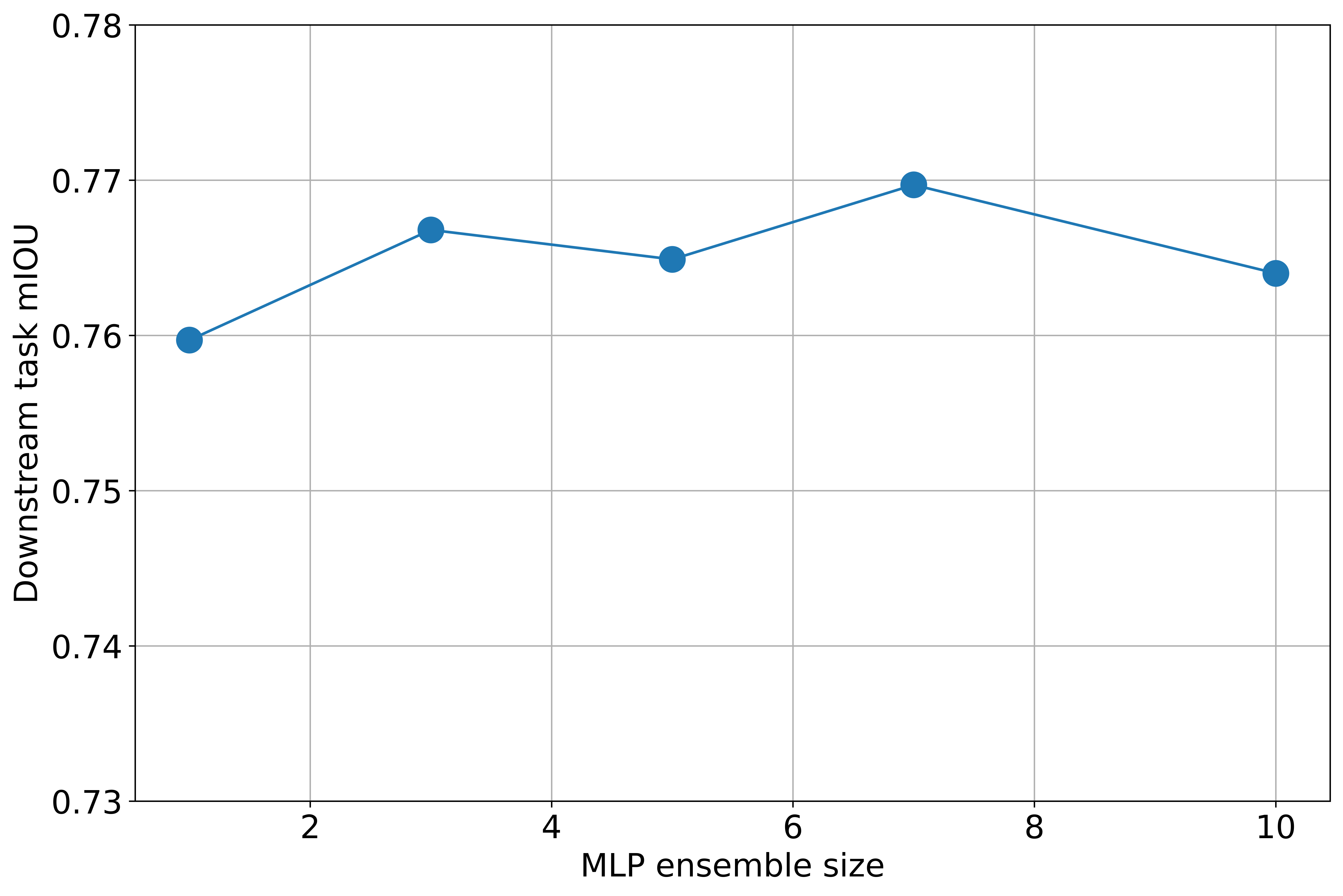}}
\hfill
\subcaptionbox{\small{Ablation for number of labeled training images in the car domain.}\label{fig:num_train_imgs}}[0.31\textwidth]
{\includegraphics[width=\linewidth]{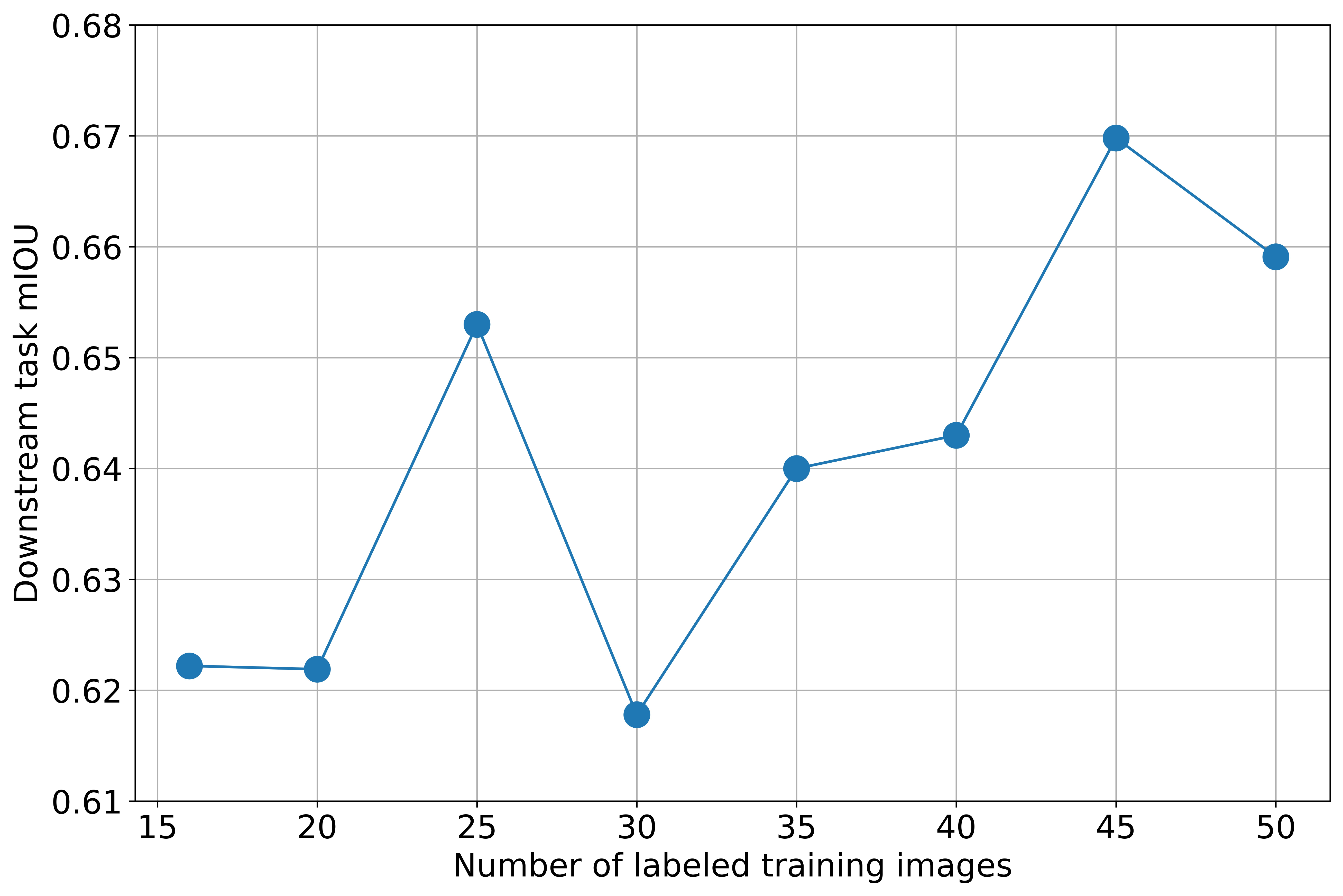}}
\hfill

\bigskip
\hfill
\subcaptionbox{\small{Ablation for the size of generated dataset in the car domain.}\label{fig:gen_data_size}}
[0.31\textwidth]
{\includegraphics[width=\linewidth]{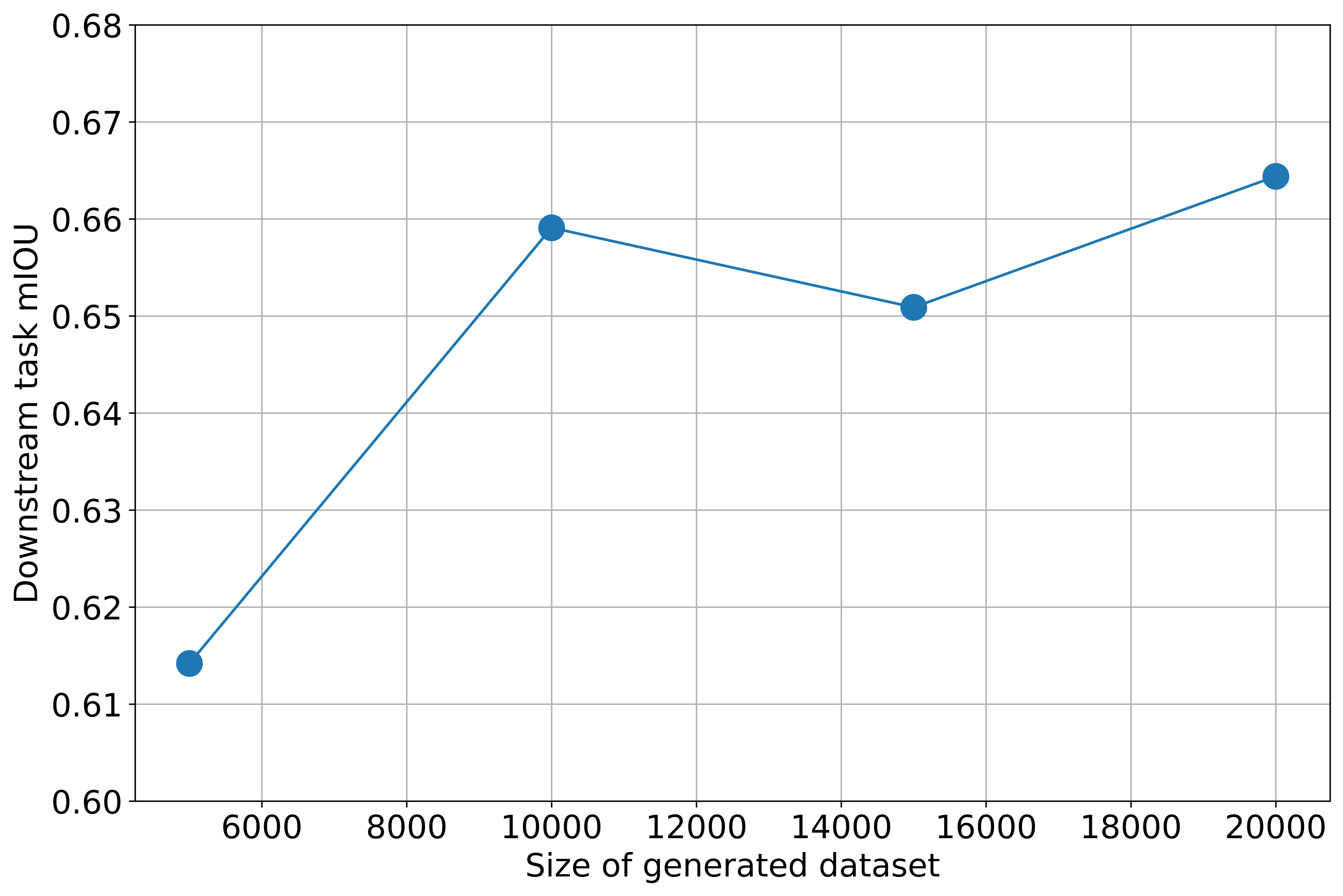}}
\hfill
\subcaptionbox{\small{Ablation for the percent of generated dataset that is filtered in the car domain.}\label{fig:filter}}[0.31\textwidth]
{\includegraphics[width=\linewidth]{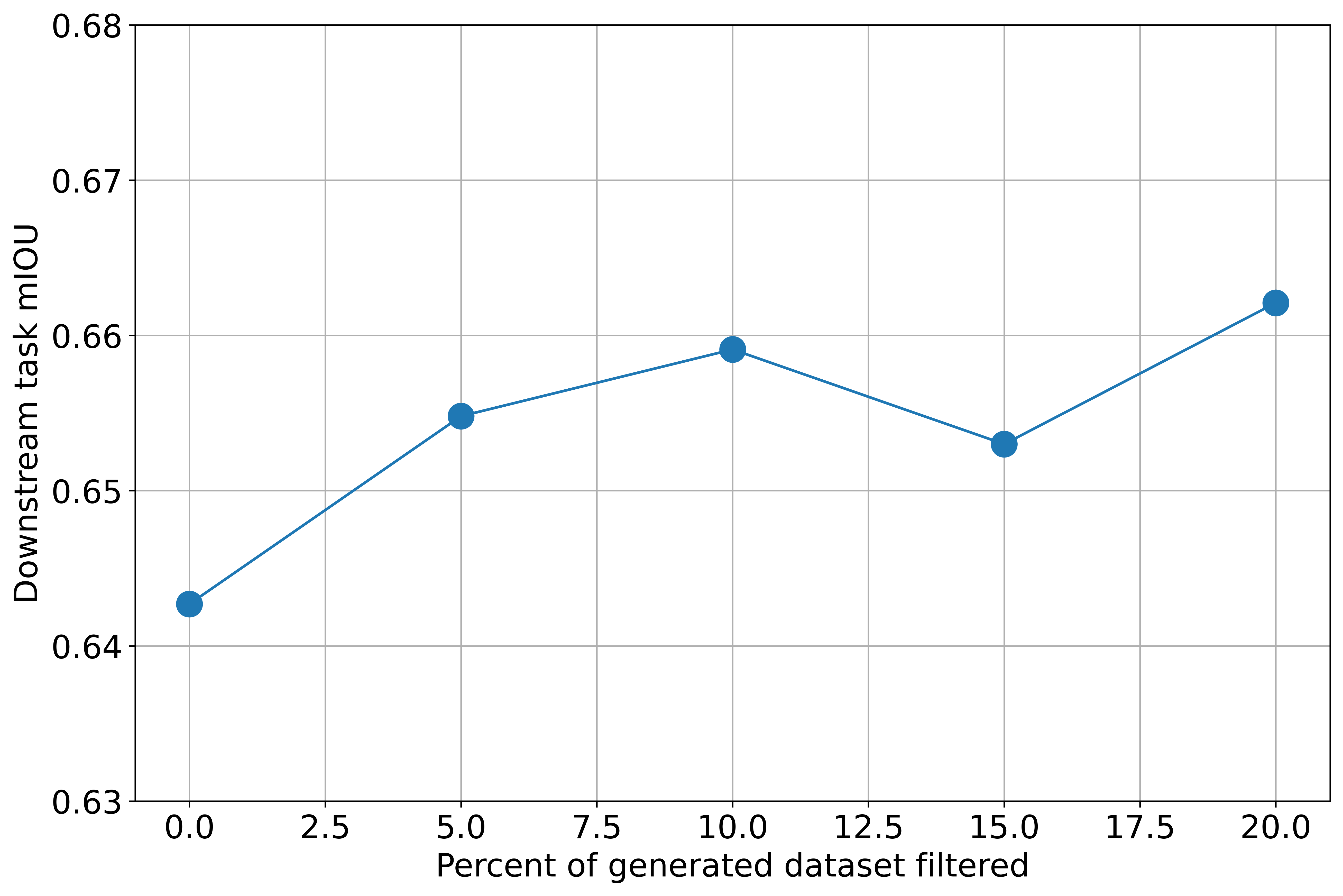}}
\hfill
\vspace{-3mm}
\end{figure*}

\paragraph{Hypercolumn dimension}  We experiment with keeping only a subset of the channels from the style block intermediate outputs from the lower resolution layers. In the StyleGAN2 generator, the first 10 style block outputs (which range from 4$\times$4 to 128$\times$128 resolutions) each contain 512 channels, comprising 5120 of the 6080 total channels. We quantify the effect of keeping zero or the first 64, 128, and 256 channels on the downstream task performance in the face domain. As shown in Tab.~\ref{fig:hypercolumn_dim}, in the face domain, while utilizing only higher resolution layers degrades performance considerably, we can remove 256 of the 512 channels for the first 10 style blocks with very minimal loss in performance. This results in a hypercolumn dimension 3520, which is a 42\% reduction compared to the original dimension of 6080. In our experiments, we utilize the full hypercolumn dimension, but note that due to memory considerations, utilizing a subset of the dimensions is feasible from a performance trade-off perspective.

\paragraph{Number of MLPs in label generator ensemble} We experiment with the number of MLPs in the ensemble. We train 1, 3, 5, 7, and 10 MLPs to generate labels. As seen in Fig.~\ref{fig:ensemble_size}, in the face domain, using only 1 network results in a performance drop, but using anywhere from 3 to 7 MLPs results in performance meeting or even exceeding the performance of using all 10 MLPs. In our experiments, we utilize 10 networks to provide for more robustness in more difficult domains, such as full-body humans and urban driving scenes.

\paragraph{Size of MLPs in label generator ensemble} We investigate whether network layer widths impact downstream performance. The original DatasetGAN framework utilizes 3-layer MLPs with intermediate dimensions of $128$ and $32$. We explore 7 additional combinations of layer widths: (256, 32), (256, 64), (256, 128), (512, 32), (512, 64), (512, 128), and (512, 256). As seen in Fig.~\ref{tab:layer_size}, in the face domain, for the face domain, downstream performance does not necessarily increase with increasing network widths, but remains relatively stable.

\paragraph{Number of labeled training images} We characterize the effects of the number of labeled training images has on downstream task performance in the car domain. As emphasized throughout the paper, a notable benefit HandsOff has over comparable frameworks is the ability for practitioners to increase the number of labeled training images without incurring costs of manual annotations. As observed in Fig.~\ref{fig:num_train_imgs}, in the car domain, the downstream performance generally increases as the number of training images is increased, but this increase is not non-decreasing. One explanation for why is that the \textit{composition} of the training data may have a larger impact on downstream performance than simply the number of images. This fact is explored in the long-tail experiments of the main paper. In our experiments, we report the performance with 16 labeled training images, which is the same number of training images in comparable baselines. We also report the performance of 50 labeled training images to highlight our framework's ability to accommodate more than a $3\times$ increase in training data.

\paragraph{Reconstruction quality} We examine the effects of GAN inversion reconstruction quality on downstream performance. Specifically, we vary the number of optimization refinement steps on the ReStyle-produced latent code. To quantitatively assess reconstruction quality, we use the value of the loss in the refinement step. As seen in Tab.~\ref{tab:recon_quality}, in the car domain, as the number of optimization iterations increases, the downstream performance generally increases. However, this increase does not scale directly with reconstruction loss.

\paragraph{Size of generated dataset} We characterize the effects of the size of the generated dataset on downstream performance. For each generated dataset size, we filter out the top 10\% uncertain images. As seen in Fig.~\ref{fig:gen_data_size}, in the car domain, as the size of the dataset grows, the downstream performance generally increases. However, the performance improvement has diminishing returns, as performance improvement is most notable moving from 5,000 to 10,000 generated image-label pairs. As a result, in our experiments, we utilize dataset sizes of 10,000 to strike a balance between performance and time and computation needed to generate larger datasets.

\paragraph{Percent of generated dataset filtered} We experiment with the percent of the dataset that is filtered out. To do so, we generate a dataset of size $10,000$ and then filter out varying percentages. As seen in Fig.~\ref{fig:filter}, in the car domain, employing filtering results in relatively similar performances. Therefore, in our experiments, we utilize a filtering percentage of 10\% to strike a balance between removing highly uncertain labels and the number of image-label pairs that are used to train the downstream model.

\paragraph{Cityscapes downstream network finetuning.} We report the effects of finetuning the trained downstream model with the original 16 or 50 labeled images used to train the label generator. As seen in Tab.~\ref{tab:cityscapes_finetune}, finetuning results in increases in performance, indicating that finetuning overcomes the difficulty in producing high quality in-distribution images with a GAN. 

\paragraph{Transfer learning pretrain dataset choice.} We report the performance of the transfer learning baseline in the face and car domain when pretrained on ImageNet versus pretrained on ImageNet and COCO. As seen in Tab.~\ref{tab:coco_pretrain}, pretraining on COCO in addition to ImageNet results in mild performance gains.

\section{Additional results}\label{sec:additional_results}
\subsection{Reconstructed image alignment}\label{sec:add_results:recon_align}
An underlying assumption of the HandsOff framework is that the reconstructed images resulting from GAN inversion align well semantically with the original labels. In this section, we present visual examples of reconstructed image alignment with original labels. 

\begin{figure*}[t]
\centering
\subcaptionbox{\label{fig:face_recon}}[\textwidth]{\includegraphics[width=\linewidth]{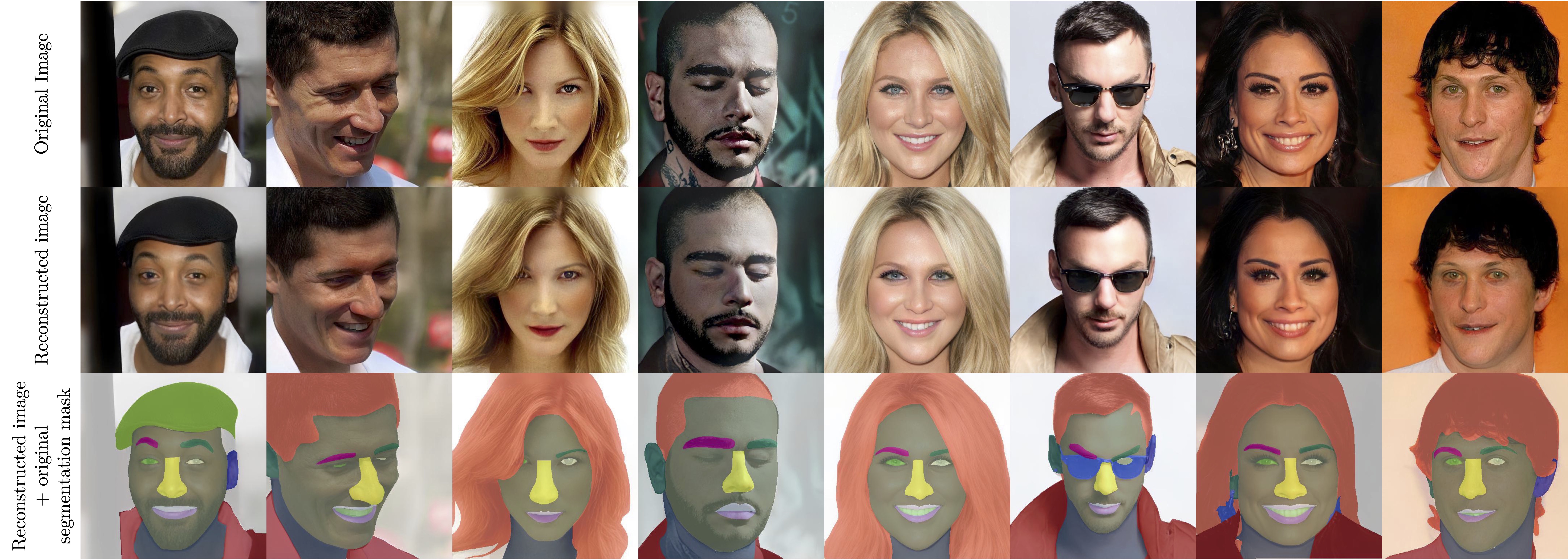}}
\vfill
\subcaptionbox{\label{fig:car_recon}}[\textwidth]{\includegraphics[width=\linewidth]{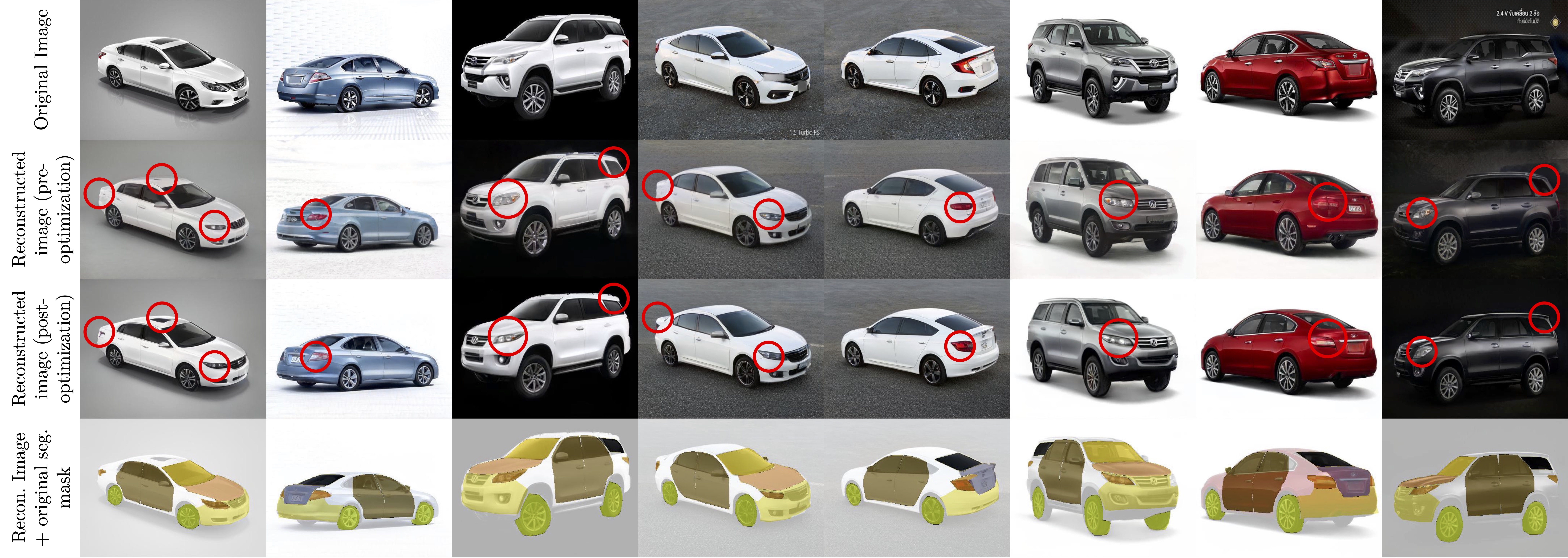}}
\caption{\small{(a) Alignment of reconstructed images with original segmentation masks in the face domain. Semantic features align almost perfectly with segmentation masks. (b) Visualization of fine detail improvement after optimization refinement in car domain. Areas of vast improvement circled in red.}}
\vspace{-3mm}
\end{figure*}

In the face domain, we utilize ReStyle for the encoder initialization and use $500$ steps of optimization to refine the images. As seen in Fig.~\ref{fig:face_recon}, the reconstructions align very well with the semantic segmentation masks from CelebAMask-HQ.

In the car domain, we utilize ReStyle for the encoder initialization and use $300$ steps of optimization to refine the images. As seen in Fig.~\ref{fig:car_recon}, the output of the ReStyle captures the overall scene very well, but struggles in preserving fine details, as shown in red circles. By utilizing the optimization based refinement step, we are able to correct for these small details. These refined images align much better with the original segmentation masks, as shown in Fig~\ref{fig:car_recon}. 

\subsection{Face domain few-shot segmentation results}\label{sec:add_results:ddpm}
In this section, we compare the downstream few-shot segmentation performance of HandsOff against self-supervised approaches and diffusion-model based approaches. Namely, we compare against DDPM-Segment \cite{baranchuk2022label}, DatasetDDPM\cite{baranchuk2022label}, MAE\cite{he2022masked}, and SwAV\cite{caron2020unsupervised}. 

DatasetDDPM and DDPM-Segment both utilize denoising diffusion probabilistic models (DDPMs). DDPM-Segment extracts intermediate network outputs from various time steps of the denoising process to form pixel-level image representations, akin to the hypercolumn representations formed from StyleGAN2 in HandsOff. Then, an ensemble of linear classifiers is trained to output a pixel-level label. DDPM-Segment is different from HandsOff in that it does not generate synthetic datasets. Instead, at inference time, the ensemble of linear classifiers is applied to the pixel-level representation of an image. DatasetDDPM simply replaces the GAN in DatasetGAN with a DDPM, forming pixel-level representations in the same manner as DDPM-Segment. For MAE and SwAV, we utilize the approach of \cite{baranchuk2022label} and extract intermediate layer outputs to form image representations of real images. We then train a segmenter to map from these representations to label outputs. 

\begin{table}[H]
    \centering
    \begin{tabular}{l c c}
        \toprule
         & \begin{tabular}{@{}c@{}} \# labeled \\ images\end{tabular} & \begin{tabular}{@{}c@{}}CelebAMask-HQ \\ 8 classes\end{tabular} \\ 
         \midrule
         DDPM-Segment & 16 & 0.772 \\
         DatasetDDPM & 20 & 0.739 \\
         MAE & 16 & 0.772\\
         SwAV & 16 & 0.725\\
         HandsOff & 16 & \textbf{0.781} \\
         \midrule
         & \begin{tabular}{@{}c@{}} \# labeled \\ images\end{tabular} & \begin{tabular}{@{}c@{}}CelebAMask-HQ \\ 19 classes\end{tabular} \\
         \midrule
         DDPM-Segment & 20 & \textbf{0.599} \\
         MAE & 20 & 0.578 \\ 
         SwAV & 20 & 0.524\\
         HandsOff & 20 & 0.583 \\
         \bottomrule
    \end{tabular}
    \caption{Segmentation task performance in face domain, reported in mIOU ($\uparrow$). Top half: experiments performed on our splits with 8 classes. Bottom half: experiments performed on \cite{baranchuk2022label} splits with 19 classes. Results for DDPM-Segment, MAE, and SwAV are those as reported in Table 2 in \cite{baranchuk2022label}.}
    \label{tab:ddpm}
\end{table}

In Tab.~\ref{tab:ddpm}, we report the performance on our train/test splits with 8 classes and the train/test splits found in \cite{baranchuk2022label} with 19 classes. With our splits and 8 segmentation classes, HandsOff outperforms all baselines, including diffusion model-based approaches DDPM-Segment and DatasetDDPM. This is likely due to two reasons: 1. DDPM-Segment does not leverage the inherent ability of generative models to produce more samples whereas HandsOff produces a large dataset on which the downstream segmenter is trained. The volume of downstream training data compensates for the advantage that diffusion models have over GANs. 2. Unlike DatasetDDPM, HandsOff trains on annotations of real images and avoids hand annotating synthetic images, which as found by \cite{baranchuk2022label}, when used in training, generally result in poorer performance. With the train/test splits found in \cite{baranchuk2022label} and 19 classes, DDPM-Segment performs slightly worse than DDPM-Segment, but outperforms the strongest self-supervised baselines (MAE \cite{he2022masked} and SwAV \cite{caron2020unsupervised}), as reported in \cite{baranchuk2022label}. We utilize the implementation of \cite{baranchuk2022label} to train DDPM-Segment end-to-end on our train/test splits. Furthermore, we utilize the publicly released synthetically generated datasets from DatasetDDPM to train a downstream network and evaluate on our train/test splits, as the labeled DDPM-generated images used to train DatasetDDPM were not publicly available.

\subsection{Additional examples of generated labels}\label{sec:add_results:label_examples}
In this section, we present additional visual examples of generated images and their labels as well as examples of segmentation mask improvements in the long-tail segmentation setting.

\begin{figure*}[t]
    \centering
    \includegraphics[height=0.92\textheight]{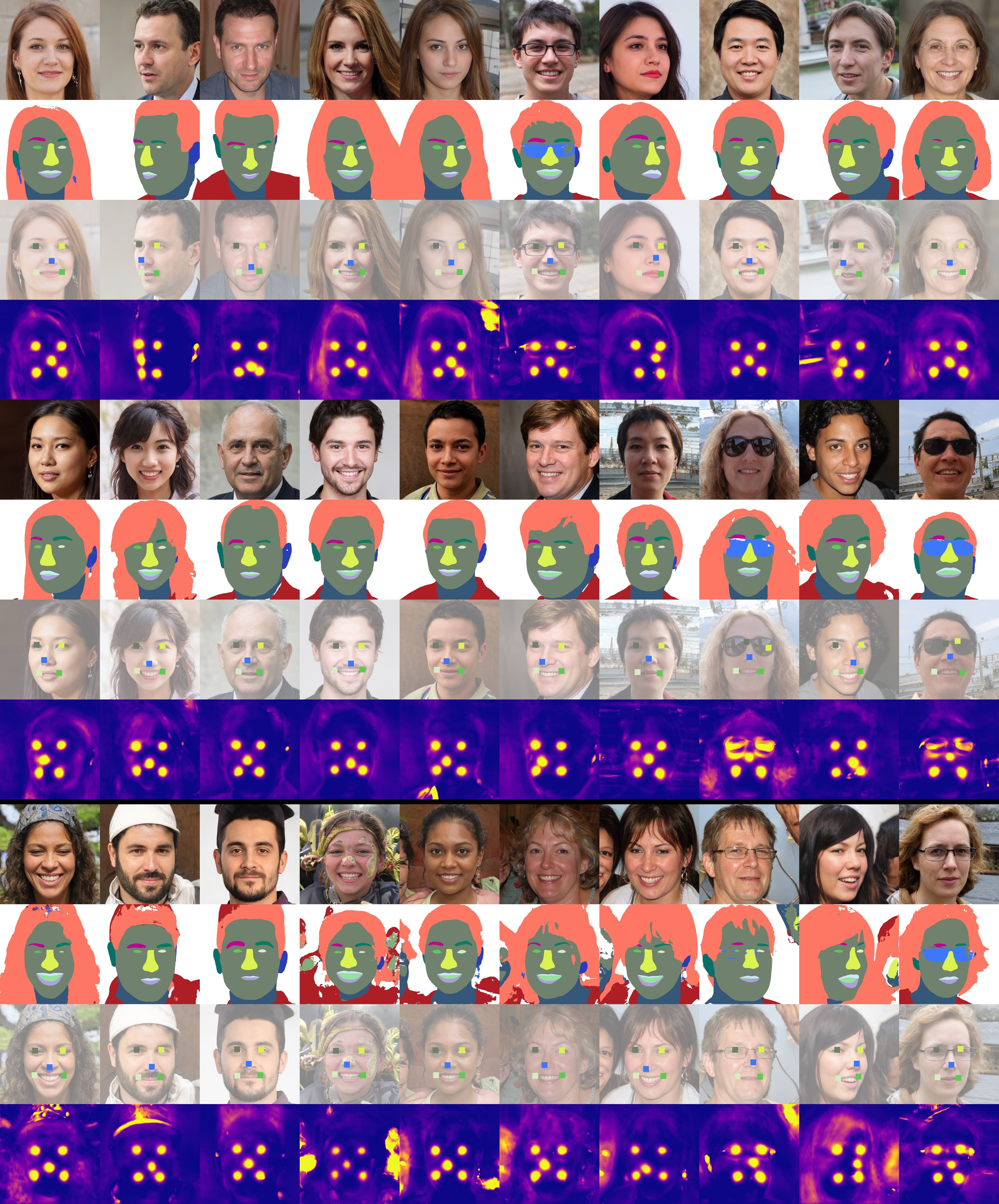}
    \caption{\small Examples of HandsOff generated labels (segmentation masks, keypoint heatmaps, and keypoints) in the face domain. Last row of examples represent typical failure cases. Hats, a rare class, are occasionally mis-classified as hair or clothing. Additionally, when the image includes GAN generated artifacts, segmentation mask quality is typically lower, while keypoint locations remain accurate.}
    \label{fig:supp_faces_gen}
    \vspace{-4mm}
\end{figure*}

\begin{figure*}[t]
    \centering
    \includegraphics[height=0.92\textheight]{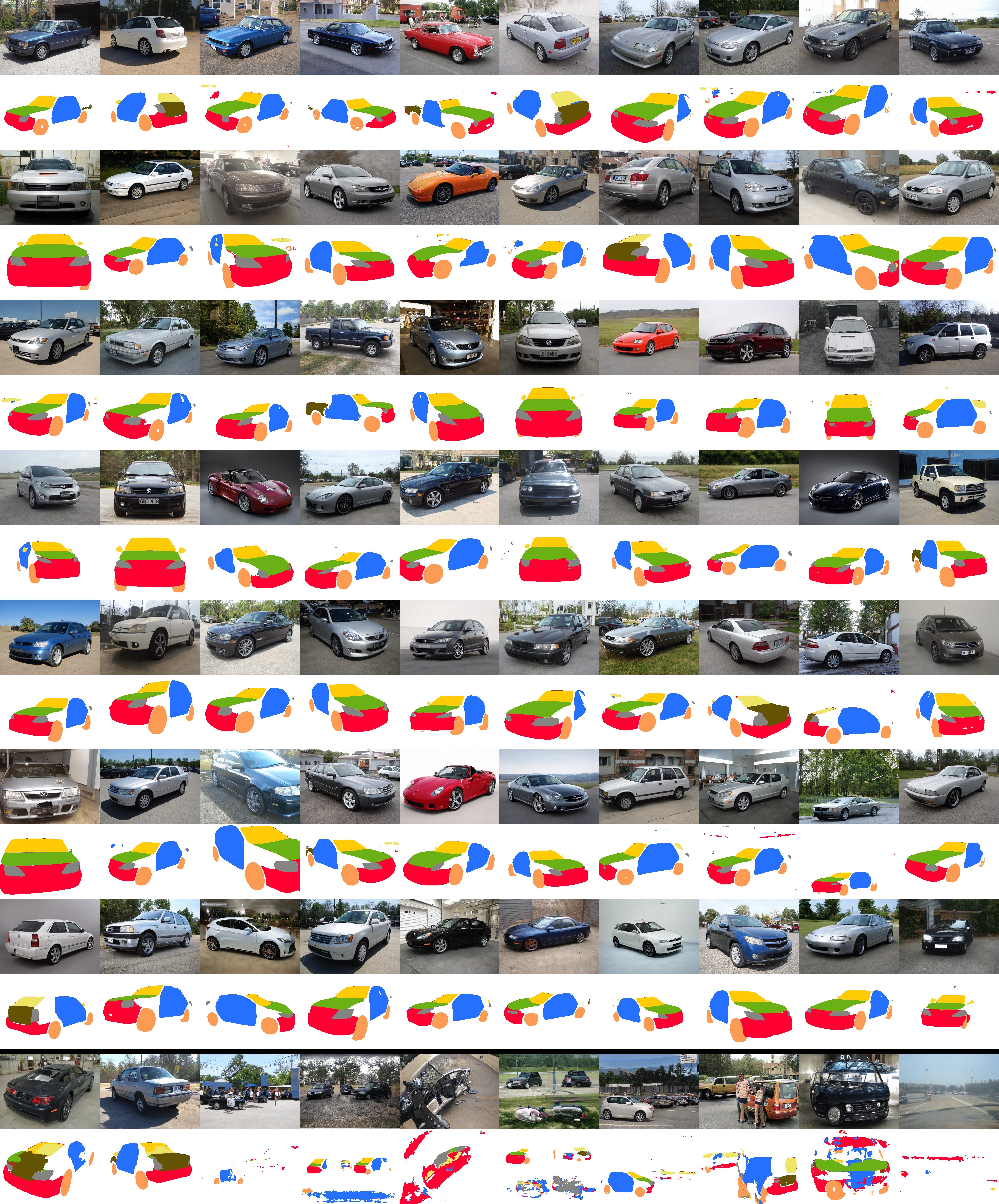}
    \caption{\small Examples of HandsOff generated segmentation masks in the car domain. Last row of examples represent typical failure cases. Similar classes, such as back trunk and front hood or front glass and back glass are confounded. Additionally, segmentation performance is typically lower when GAN generated images are out of domain or incoherent.}
    \label{fig:supp_car_gen}
    \vspace{-4mm}
\end{figure*}

\begin{figure*}[t]
    \centering
    \includegraphics[height=0.92\textheight]{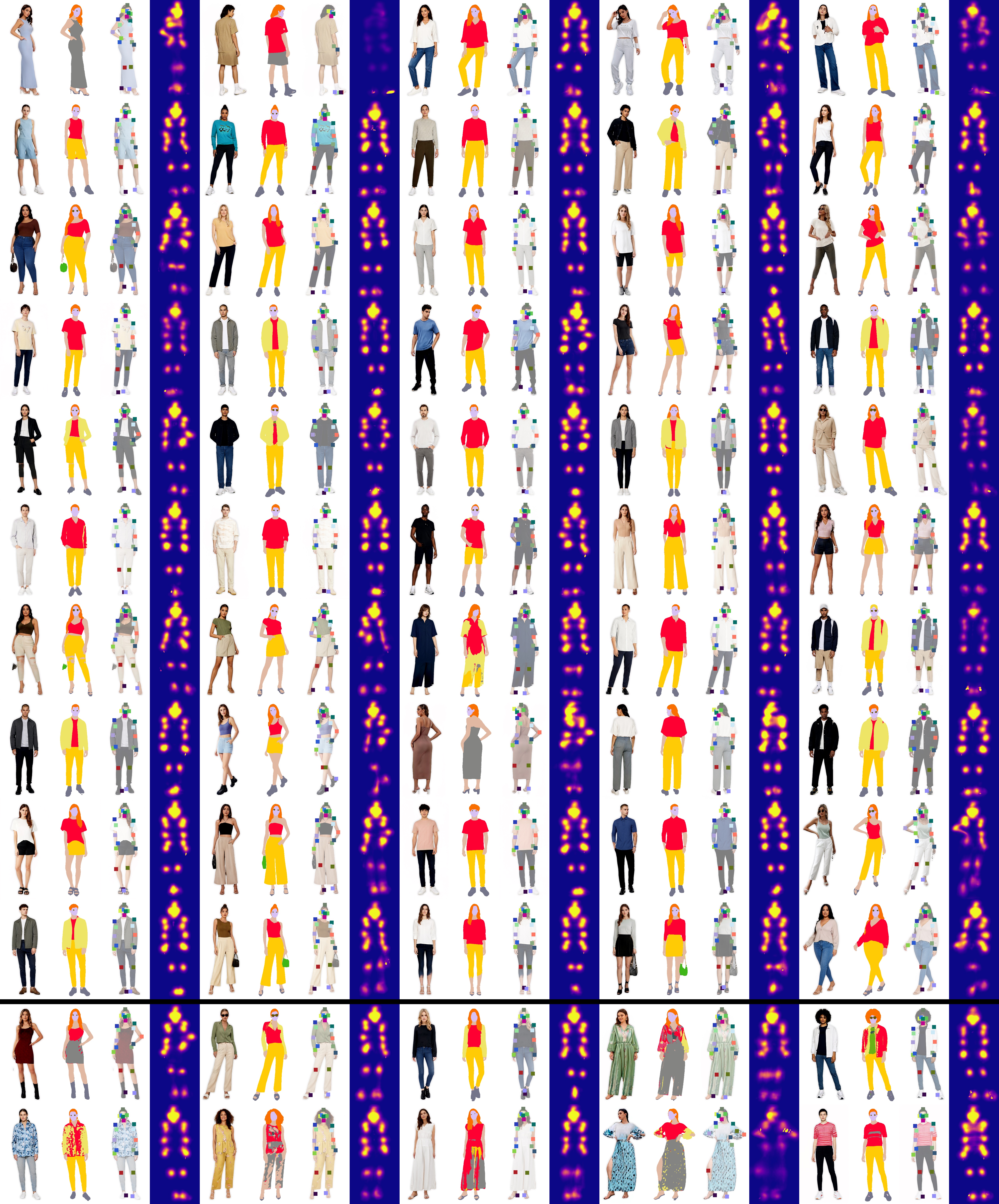}
    \caption{\small Examples of HandsOff generated labels (segmentation masks, keypoint heatmaps, and keypoints) in the full-body human poses domain. Last row of examples represent typical failure cases. Similar classes, tops, outerwear, and dresses are confounded. Furthermore, patterned pieces of clothing seem to result in mixed segmentation performance. Keypoint locations remain accurate even when segmentation masks are of lower quality.}
    \label{fig:supp_fashion_gen}
    \vspace{-4mm}
\end{figure*}

\begin{figure*}[t]
    \centering
    \includegraphics[height=0.92\textheight]{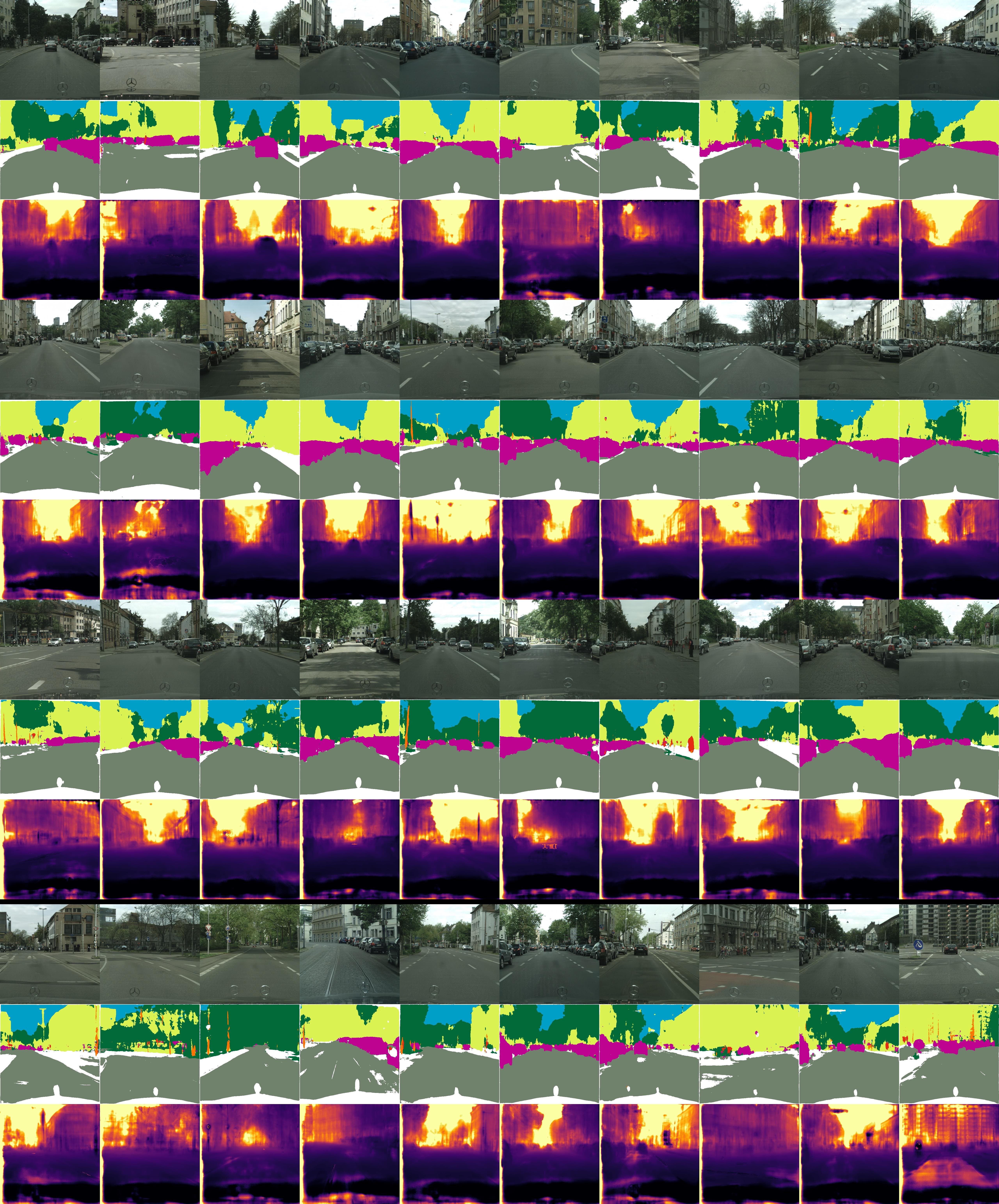}
    \caption{\small Examples of HandsOff generated labels (segmentation masks and depth maps) in the urban driving scenes domain. Last row of examples represent typical failure cases. Visually small objects such as light poles and street signs are often confounded as background classes or not labeled. In cases of background buildings with many vertical lines, such lines can be mistaken as street sign poles (last image in last row). Depth maps remain relatively accurate even when segmentation masks are of lower quality.}
    \label{fig:supp_city_gen}
    \vspace{-4mm}
\end{figure*}

\begin{enumerate}
    \item In Fig.~\ref{fig:supp_faces_gen}, we present examples in the face domain. We include examples of the predicted aggregated keypoint heatmaps used to generate the predicted keypoints. To produce the aggregated heatmap, we sum across all of the individual keypoint heatmaps.
    \item In Fig.~\ref{fig:supp_car_gen}, we present examples in the car domain.
    \item In Fig.~\ref{fig:supp_fashion_gen}, we present examples in the full-body human pose domain. We again include examples of aggregated predicted heatmaps used to generate the predicted keypoints. To produce the aggregated heatmap, we sum across all of the individual keypoint heatmaps.
    \item In Fig.~\ref{fig:supp_city_gen}, we present examples in the urban driving scene domain.
\end{enumerate}

\subsection{Additional examples of long-tail visualizations}\label{sec:add_results:longtail}
\begin{figure*}[t]
\centering
\subcaptionbox{\label{fig:glasses_longtail}}[0.9\textwidth]{\includegraphics[width=0.9\linewidth]{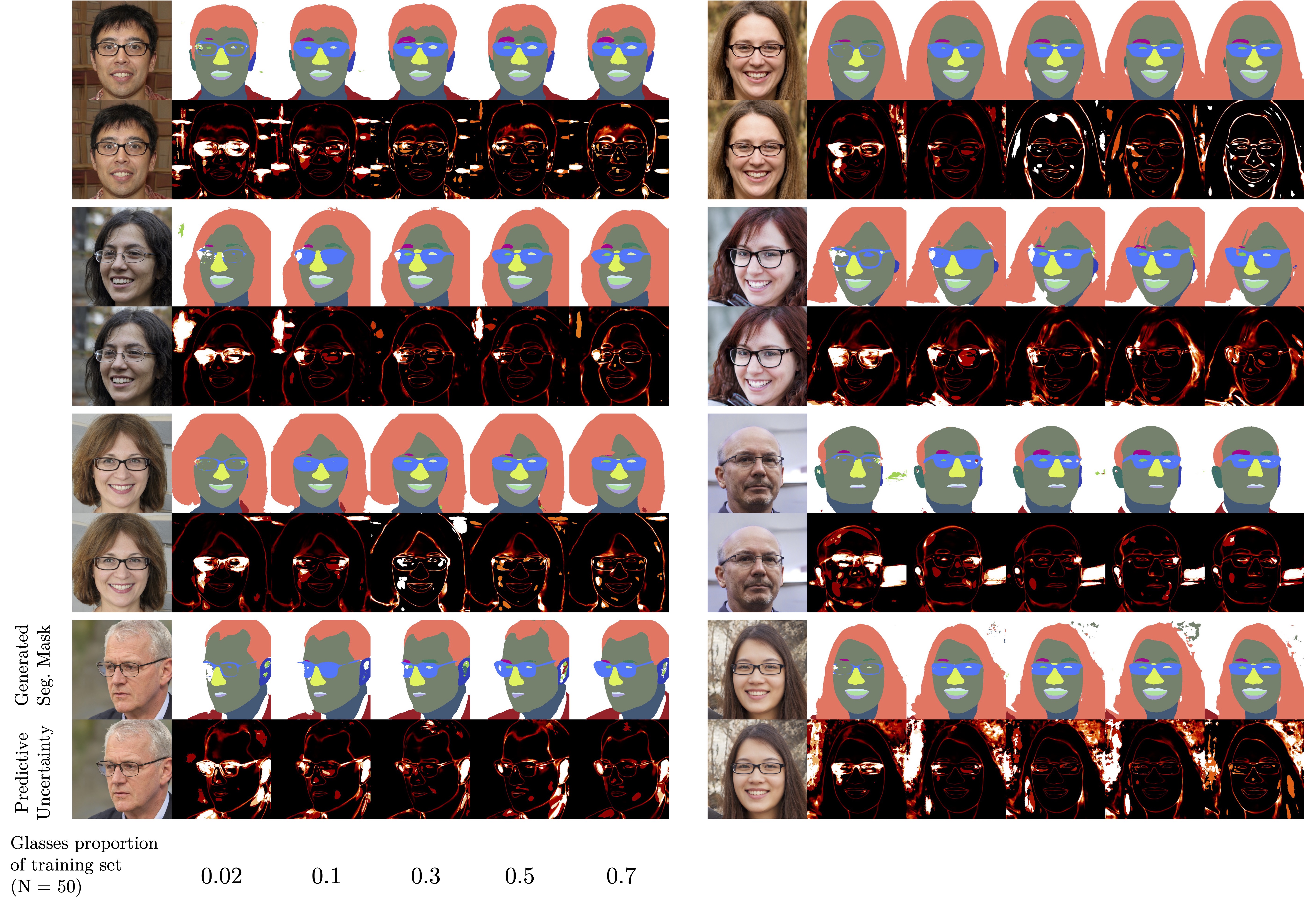}}
\vfill
\subcaptionbox{\label{fig:hat_longtail}}[0.9\textwidth]{\includegraphics[width=0.9\linewidth]{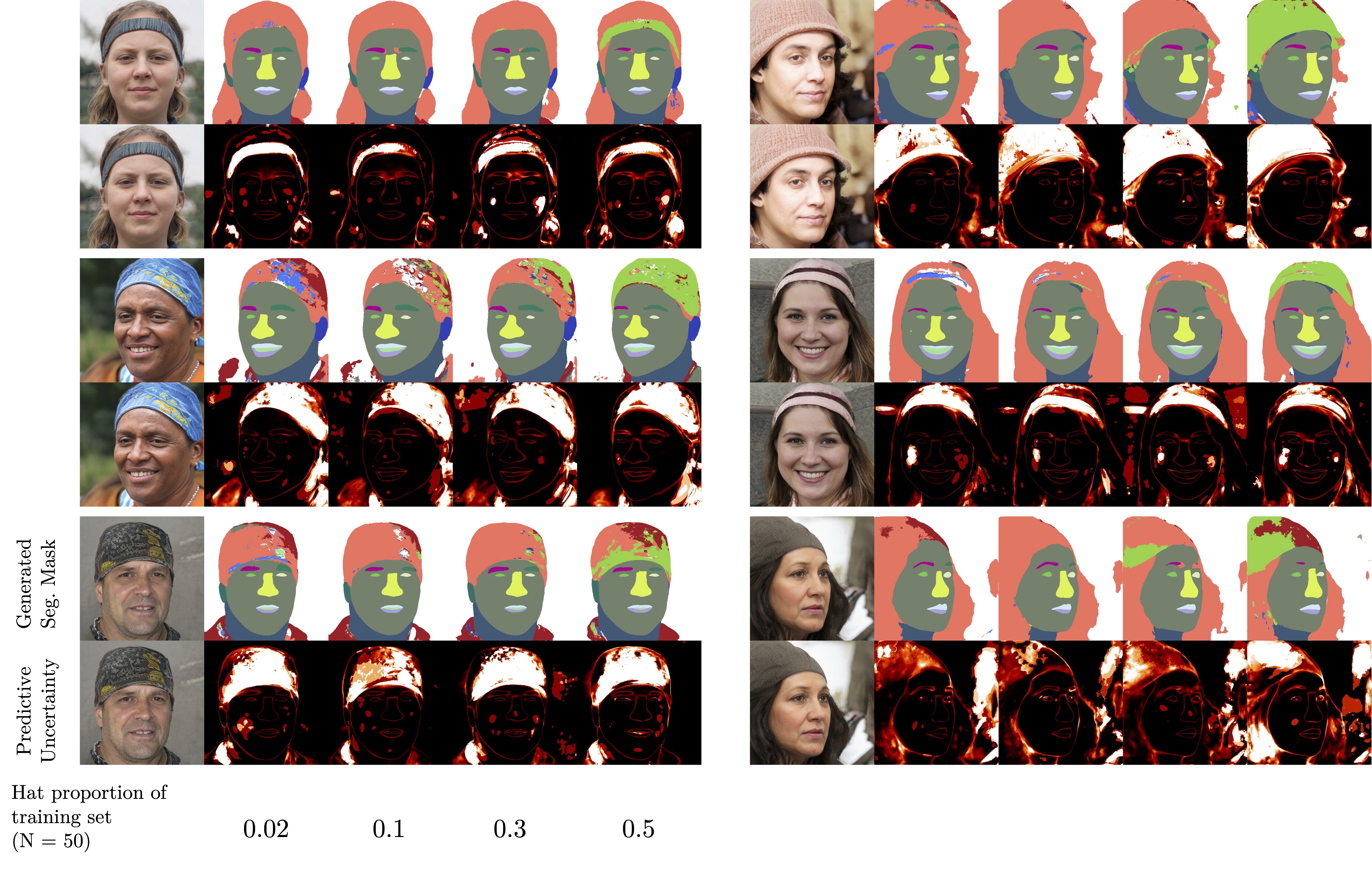}}
\caption{\small{Visualization of generated segmentation mask and pixel-wise label generator uncertainty. (a) Not only do we see qualitative improvement in the generated label for glasses, we also
see that the classifier is less uncertain when generating the correct label. (b) Hats are a particularly challenging class to characterize, so while the quality of the masks improves drastically, the classifier uncertainty remains relatively high. The last row of examples shows typical failure cases, where the hat is classified as semantically similar classes, such as hair or clothing.}}
\vspace{-3mm}
\end{figure*}

In Fig.~\ref{fig:glasses_longtail} and~\ref{fig:hat_longtail}, we present examples of long-tail segmentation mask progressions and pixel-wise uncertainty measurements with glasses and hats, respectively. Uncertainty is measured by Jensen-Shannon divergence (See Sec.~\ref{sec:framework:label_gen}).

\begin{table*}[t]
    \begin{tabular}{l | c c c c c c c c}
        MLP layer widths & (128, 32) & (256, 32) & (256, 64) & (256, 128) & (512, 32) & (512, 64) & (512, 128) & (512, 256) \\
        \noalign{\hrule height 1pt}
        mIOU & 0.7740  & \textbf{0.7859}  & 0.7813 & 0.7807 & 0.7828 & 0.7818 & 0.7817 & 0.7850\\ 
    \end{tabular}
    \caption{\small{Ablation for MLP hidden layer widths in the face domain}}\label{tab:layer_size}
    \vspace{6mm}
    
    \begin{tabular}{l | c c c c c c}
        Optimization loss  & 3.333 & 2.292 & 2.185 & 2.140 & 2.108 & 2.089 \\
        Optimization iterations & 0 & 100 & 200 & 300 & 400 & 500\\
        \noalign{\hrule height 1pt}
        mIOU & 0.5735 & 0.6278 & 0.6301 & \textbf{0.6679} & 0.6426 & 0.6591
    \end{tabular}
    \caption{\small{Ablation for GAN inversion quality in the car domain.}}\label{tab:recon_quality}
    \vspace{6mm}

    \begin{tabular}{c c c}
          \# labeled images & No finetuning & Finetuning  \\
          \noalign{\hrule height 1pt}
          16 & 0.5206 & 0.5510\\
          50 & 0.5492 & 0.6047
    \end{tabular}
    \caption{\small{Ablation for Cityscapes downstream network finetuning.}}\label{tab:cityscapes_finetune}
    \vspace{6mm}

    \begin{tabular}{l c c c}
         Domain & \# labeled images & ImageNet pretrain & COCO + ImageNet pretrain  \\
          \noalign{\hrule height 1pt}
         Faces & 16 & 0.4575 & 0.4896\\
         Faces & 50 & 0.6197 & 0.6295\\
         Cars & 16 &  0.3232 & 0.3313\\
         Cars & 50 &  0.4802 & 0.5026 
    \end{tabular}
    \caption{\small{Ablation for choice of pretraining dataset for transfer learning baseline.}}\label{tab:coco_pretrain}
\end{table*}

\end{document}